\documentclass[12pt,maitrise,oneside]{dms} %Remplacer <maitrise> par <phd> pour une thèse

\usepackage[utf8]{inputenc} %Obligatoires
\usepackage[T1]{fontenc}    %

\usepackage[longnamesfirst,numbers]{natbib}
\usepackage[french,english]{babel}
\anglais

\def\sloppy{%
  \tolerance 500%  %9999 dans LaTeX ordinaire, mauvaise idée.
  \emergencystretch 3em%
  \hfuzz .5pt
  \vfuzz\hfuzz}
\usepackage{graphicx,amssymb,subfigure,icomma}
\usepackage[labelfont=bf, width=\linewidth]{caption}

\usepackage{hyperref}  % Ajoute les hyperlien
\hypersetup{colorlinks=true,allcolors=black}
\usepackage{hypcap}   % Corrige la position du lien pour les images
\usepackage{bookmark} % Remédie à des petits problème
\usepackage{algorithm}
\usepackage{algpseudocode}
\usepackage{tikz}

\numberwithin{equation}{section}
\numberwithin{table}{chapter}
\numberwithin{figure}{chapter}

\begin{document}

% La commande "\brouillon" imprime, au bas de chaque page, la date ainsi que l'heure de la derni\`ere compilation de votre fichier.
% \brouillon

% Voici les variables pour la création de votre page titre.

\title{Analyzing the Benefits of Communication Channels Between Deep Learning Models}
\author{Philippe Lacaille}
\copyrightyear{2018}
\date{Août 2018}									% Date de dépôt du document.
	% ces éléments ne doivent plus apparaittre selon les dierectives de la FESP
	% si toutefois vou souhaitez les inclure, il faudra aussi modifier le document dms.cls
% \president{Nom du président du jury}
% \directeur{Yoshua Bengio}
% \codirecteur{Nom du codirecteur}
% \membrejury{Nom du membre du jury} 
% \examinateur{Nom de l'examinateur externe}
% %\membresjury{alpha, beta, gamma}
% %\plusmembresjury{psi, zeta, omega} 
% \repdoyen{Nom du représentant du doyen} 
\dateacceptation{Date d'acceptation}
\sujet{informatique}							% Votre discipline de recherche, soit "mathématiques" ou "statistique".
%\orientation{mathématiques fondamentales}		% Cette commande est optionnelle. Les choix courants sont : "mathématiques fondamentales", "mathématiques de l'ingénieur" et "mathématiques appliquées".

\department{Département d'informatique et de recherche opérationnelle}

% Fin des variables \`a définir. La commande "\maketitle" créera votre page titre.

\pagenumbering{roman}
\maketitle
\begin{otherlanguage}{french}
\chapter*{Sommaire} 	% La commande "\chapter*" crée un chapitre sans numéro, contrairement \`a la commande "\chapter" réguli\`ere.
% Sommaire

\noindent
Comme les domaines d'application des systèmes d’intelligence artificielle ainsi que les tâches associées ne cessent de se diversifier, les algorithmes d'apprentissage automatique et en particulier les modèles d'apprentissage profond et les bases de données requises au fonctionnement de ces derniers grossissent continuellement. Certains algorithmes permettent de mettre à l'échelle les nombreux calculs en sollicitant la parallélisation des données. Par contre, ces algorithmes requièrent qu'une grande quantité de données soit échangée afin de s'assurer que les connaissances partagées entre les cellules de calculs soient précises.

Dans les travaux suivants, différents niveaux de communication entre des modèles d'apprentissage profond sont étudiés, en particulier l'effet sur la performance de ceux-ci. La première approche présentée se concentre sur la décentralisation des multiples calculs faits en parallèle avec les algorithmes du gradient stochastique synchrone ou asynchrone. Il s'avère qu'une communication simplifiée qui consiste à permettre aux modèles d'échanger des sorties à petite bande passante peut se montrer bénéfique. Dans le chapitre suivant, le protocole de communication est modifié légèrement afin d'y communiquer des instructions pour l'entraînement. En effet, cela est étudié dans un environnement simplifié où un modèle préentraîné, tel un professeur, peut personnaliser l'entraînement d'un modèle initialisé aléatoirement afin d'accélérer l'apprentissage. Finalement, une voie de communication où deux modèles d'apprentissage profond peuvent s'échanger un langage spécifiquement fabriqué est analysée tout en lui permettant d'être optimisé de différentes manières.
\\

\noindent \textbf{Mots-clés:} Apprentissage automatique, apprentissage profond, communication, langage, professeur, étudiant, optimisation, gradient
\end{otherlanguage}
\chapter*{Summary}
% Summary

\noindent
As artificial intelligence systems spread to more diverse and larger tasks in many domains, the machine learning algorithms, and in particular the deep learning models and the databases required to train them are getting bigger themselves. Some algorithms do allow for some scaling of large computations by leveraging data parallelism. However, they often require a large amount of data to be exchanged in order to ensure the shared knowledge throughout the compute nodes is accurate.

In this work, the effect of different levels of communications between deep learning models is studied, in particular how it affects performance. The first approach studied looks at decentralizing the numerous computations that are done in parallel in training procedures such as synchronous and asynchronous stochastic gradient descent. In this setting, a simplified communication that consists of exchanging low bandwidth outputs between compute nodes can be beneficial. In the following chapter, the communication protocol is slightly modified to further include training instructions. Indeed, this is studied in a simplified setup where a pre-trained model, analogous to a teacher, can customize a randomly initialized model's training procedure to accelerate learning. Finally, a communication channel where two deep learning models can exchange a purposefully crafted language is explored while allowing for different ways of optimizing that language.
\\

\noindent
\textbf{Keywords:} Machine learning, deep learning, communication, language, teacher, student, optimization, gradients

% TABLE DES MATIÈRES
\cleardoublepage
\pdfbookmark[chapter]{\contentsname}{toc}  %Crée un bouton sur la bar de navigation
\tableofcontents				% Table des mati\`eres.
% LISTE DES TABLEAUX
\cleardoublepage
\phantomsection
\listoftables
% LISTE DES FIGURES
\cleardoublepage
\phantomsection
\listoffigures

%%%%%%%%%%%%%%%%%%%%%%%%%%%%%%%%%%%%%
%% LISTE DES SIGLES ET ABRÉVIATION %
%%%%%%%%%%%%%%%%%%%%%%%%%%%%%%%%%%%%%
%% Il est obligatoire, selon les directives de la FESP, 
%% pour une thèse ou un mémoire d'avoir une liste des sigles et 
%% des abréviations.  Si vous considérez que de telles listes ne seraient pas
%% pertinentes (si, par exemple, vous n'utilisez aucun sigle ou abré.), son
%% inclusion ou omission est laissé à votre discrétion.  En cas de doute,
%% parlez-en à votre directeur de recherche, le coadministrateur ou, ultimement, à
%% la FESP directement.
%%
%% Dans le cas où vous incluez une table des sigles et des abréviations,
%% vous pouvez enlever les % de la section suivante pour faire apparaître
%% un exemple d'une telle liste « fait à la main ».  Il existe des outils
%% plus sophistiqués si vous devez inclure une multitude de sigles et abréviations.
%% (Par exemple, le package <glossaries> peut faire des index élaborés.  Comme
%% son utilisation est technique, il n'y a pas d'exemple directement dans ce gabarit.
%% On invite les gens qui aurait à l'utiliser à consulter le wiki
%% du dms, le coadministrateur ou faire leur propre recherche.)

\chapter*{List of Abbreviations}
\begingroup %Pour que le \renewcommand soit local
%Modifiez ce nombre (p.ex.remplacez 2 par 1.5) pour augmenter ou diminuer l'espace entre les lignes du tableau.
\renewcommand{\arraystretch}{1.0} 
\noindent\begin{tabular}{p{.2\textwidth} p{.7\textwidth}}
 SGD  & Stochastic Gradient Descent \\
 GPU & Graphics Processing Unit \\
 MSE  & Mean Square Error \\
 MLP & Multi-Layer Perceptron \\
 RNN & Recurrent Neural Network \\
 CNN  & Convolutional Neural Network \\
 ReLU & Rectified Linear Unit \\
 GAN & Generative Adversarial Network \\
 VAE & Variational Auto-Encoder \\
 ELBO & Evidence Lower Bound \\
 DCGAN & Deep Convolutional Generative Adversarial Network \\
 BiGAN & Bidirectional Generative Adversarial Network \\
 KL & Kullback-Leibleir \\
 MINE & Mutual Information Neural Estimation \\
\end{tabular}
\endgroup  %Fin du groupe local pour \arraystretch

\chapter*{Acknowledgements} %Remerciements

% Acknowledgements

\noindent
Firstly, I would like to thank my supervisor Yoshua Bengio, who was willing to give me, an actuary coming back to school, a chance to learn about this field of which I am now passionate about. Your guidance and your support, both educational and financial, during the completion of this Master's degree will serve as an inspiration in the future; merci Yoshua.

To my close family, Jean-Claude, Carole, Olivier and, of course, Marie-Pier, thank you for believing in me and keeping me steered in the right direction through the ups and downs of the past few years, your support means everything to me.

To my collaborator of all the work presented in this thesis, Min Lin, I can't thank you enough for your accessibility and your patience. Through the many brainstorming sessions, along with your help writing and debugging code, I hope you enjoyed our time working together as much as I did.

Thank you also to all my colleagues and friends from room 3248 at Mila, the numerous hours spent there were truly enjoyable.

Finally, thank you to my dear friend William Perrault, who made getting back to school as enjoyable as it was. I'm forever grateful our paths crossed, to many more golf rounds and even more birdies in the future.

% Fin des pages liminaires. À partir d'ici, les premi\`eres pages des chapitres ne doivent pas être numérotées.

% Voici maintenant le chapitre d'introduction.
\NoChapterPageNumber 
\cleardoublepage

\pagenumbering{arabic}

\chapter*{Introduction}
\pagenumbering{arabic}
% Introduction

\noindent
This thesis presents my research during the completion of my Master's degree at the Université de Montréal at Mila under the supervision of Professor Yoshua Bengio. This work was done in the field of computer science and more specifically of artificial intelligence and with the collaboration of postdoctoral researcher Min Lin at Mila.

This thesis is structured in such a way to introduce the machine learning basics in the first chapter to be able to follow the work detailed in the subsequent ones. Chapters 2, 3 and 4 detail different experiments to study how useful having a communication channel between deep learning models can be.

\chapter{Machine Learning Basics}
% Machine Learning Basics

In this chapter, a review of the basics of machine learning required to follow the work described in the following chapters is made. This does not serve as a full review of machine learning nor deep learning. Should the reader be interested in an in-depth review of the background material and trends in machine learning and more specifically in deep learning, the Deep Learning book \cite{Goodfellow2016deeplearning} is an excellent alternative.

\section{Introduction}

Machine learning is the greater family of creating functions from data, leveraging the computing abilities and algorithms of computer science. In addition to computer science, it is at the intersection of multiple fields of research, in particular, probability and statistics, information theory, optimization, linear algebra and linguistics.

Recent breakthroughs in artificial intelligence such as the highly publicized Alpha Go successes \cite{silver2016mastering}, leverage different components of the machine learning family. One in particular, reinforcement learning is not covered throughout this document. The interested readers can learn more about it from \cite{sutton1998introduction}, which a second edition is currently in the works.

\section{Tasks}

Although the field of machine learning and its potential applications are expanding rapidly, most of them can be boiled down to a couple categories of tasks. This section serves as a brief introduction to the most common tasks that leverage machine learning algorithms.

\subsection{Supervised learning tasks}

Supervised learning tasks can be generally seen as tasks where the objective is to make a prediction as to what an input corresponds to. Conceptually, models try to figure out what is the relationship between the input data and an associated value or label. Generally speaking, the objective is to build models such that for similar input, it makes similar predictions in the hopes of being able to make an appropriate prediction for new data. In other words, supervised learning tasks aim to understand the relationship between the data and some other value or attribute of the data. Provided with a dataset with training samples, the difficulty lies in establishing what consists of a similarity.

\subsubsection{Classification}

A typical classification task consists of making a prediction regarding which class or group a given input associates with. Examples of such task includes predicting if an image is one of a cat or a dog, or given a wine sample predict if it is a red or white wine. Other \textit{advanced} tasks such as auto-correcting mistyped words on a cellphone keyboard or even facial recognition software all correspond to a form of classification task. Apart from some cases of multi-label classification, the general objective of the machine learning model in this task is to predict which one of the possible $M$ classes corresponds to that input.

Datasets used for classification tasks consist of the input data as well as the corresponding label for each of the data samples. The goal is then to make a prediction on the label of a new data point. To design a classification algorithm or model, it is often mandatory to know ahead of time the possible labels or classes that the data may represent, i.e. will these pictures be exclusively of either dogs or cats? In addition, the loss function used to train machine learning models for supervised learning tasks is the cross-entropy loss between the model's probabilistic prediction and the label distribution of the training data.

Given training data $D_{train} = \{(x^{(1)}, y^{(1)}), \dots, (x^{(i)}, y^{(i)}), \dots, (x^{(N)}, y^{(N)})\}$, where $x \in \mathbb{R}^d , y \in \{1, 2, \dots, C\}$ and $f_\theta (x^{(i)}) = \big[ P(y^{(i)}=1|x^{(i)}), P(y^{(i)}=2|x^{(i)}), \dots, P(y^{(i)}=C|x^{(i)}) \big]$, the global cross-entropy objective can be defined as,

\begin{equation}
L(f_\theta(x^{(i)}), y^{(i)})  = - \sum\limits_{j=1}^C \mathbf{1}_{\{y^{(i)}=j\}}\log f_\theta (x^{(i)})_j
\label{eq:cross-entropy}
\end{equation}

\subsubsection{Regression}

Similarly to classification, regression tasks can be seen as making a prediction of value for a given input. However, instead of selecting one of the possible groups to associate a given input to, the goal is to estimate a real value. An intuitive example to understand the regression is to consider estimating the value of a house. A real estate agent acts similarly to a machine learning model in a regression task where given all the information of the neighborhood and characteristics of a house, it tries to determine a good market price to list the house on the market. Variants of regression include predicting the amount of acceleration an autonomous vehicle requires or even the price a user may be ready to pay online for an item given its user profile.

The datasets for regression are similar to classification tasks but where a real value is associated with each of the data samples. The loss function generally used in regression tasks is the mean-square error (MSE), as it allows for computing losses between real numbers.

Given training data $D_{train} = \{(x^{(1)}, y^{(1)}), \dots, (x^{(i)}, y^{(i)}), \dots, (x^{(N)}, y^{(N)})\}$, where $x \in \mathbb{R}^d , y \in \mathbb{R}$ and $f_\theta (x^{(i)}) = \hat{y}^{(i)}$, the MSE objective can be defined as,

\begin{equation}
L(f_\theta(x^{(i)}), y^{(i)}) = \frac{1}{2} (y^{(i)} - f_\theta(x^{(i)}))^2
\label{eq:mean-square-error}
\end{equation}

\subsection{Unsupervised learning tasks}

Unlike supervised learning tasks, unsupervised learning focuses on the actual data itself rather than its relationship with a corresponding label or value. There is a wide array of unsupervised learning tasks, but most of them try to estimate, in some way, the underlying distribution of the data. In general, a key distinction from supervised learning datasets is the absence of labels.

\subsubsection{Clustering}

Conceptually, clustering is a task about discovering boundaries throughout the data in order to regroup similar data points into groups, or clusters. The number of clusters is usually required to be known in advance. There is an analogy to be made between clustering and classification, only the former is in a situation where it is not known in advance what classes the data correspond to. An example of clustering tasks include grouping users based on their online activity in order to better predict their purchases.

\subsubsection{Density estimation}

Rather than focus on a function of the data, density estimation aims at discovering or approximating the underlying function that is \textit{behind} the data. This function is commonly called the probability density function. In other words, the goal is to find the distribution from which the data was created. If successful, the recovered function is a powerful tool that can be used to replace the original data or even create new data from the same distribution. Density modelling can further be understood as a way of compressing all the data on hand into a machine learning model.

These tasks are pretty general and depend on what the intended purpose is once the density has been successfully modelled by a machine learning model. For example, modelling the insurance claims of a set of car insurance policies allows an insurance company to analyze the risks they are exposed to. Given that model, it can further determine an appropriate pricing for a new customer. The key here is the intent of doing something else with the density estimation but where on its own, it doesn't really do anything.

\subsubsection{Generative models}

Density estimation usually requires to explicitly have a parameterized model of a family of density functions. On the contrary, generative models usually have an implicit model of the density distribution of the data. The objective is to be able to then sample that distribution in order to get some additional data samples. An example of tasks using generative models would be to create additional artwork given some paintings of a deceased artist. By modelling the distribution of the known paintings of an artist, it could be possible, in theory, to then generate new paintings that correspond to that artist's characteristics.

\section{Optimization and evaluation}

In order to solve any of the previously mentioned tasks in machine learning, a mathematical framework of the problem needs to be formulated. Once that problem is formulated as an optimization problem, some algorithms can be used to minimize or maximize the objective. Throughout the different tasks, the one recurring approach is to define the objective to minimize as a loss function. That loss function is what defines the task a model will be trained to do.

Given a function $f$ with parameters $\theta$ and a training dataset $D_{train}$, the global objective $J(\theta)$ can be defined as,

\begin{equation}
J(\theta) = \hat{R}(f_\theta, D_{train}) = \dfrac{1}{|D_{train}|}\sum \limits_{i = 1}^{|D_{train}|} L(f_\theta(x^{(i)}), y^{(i)})
\end{equation}

The ultimate \textit{goal} of the optimization procedure is to find the parameters $\theta$ that minimize $J(\theta)$, the empirical risk, over the entire training data. The solution to the optimization problem can be written as $\theta^* = \operatorname*{argmin} \limits_\theta J(\theta)$.

However, this is not easily accomplished, in particular given the complexity of the tasks that are now explored in the machine learning community. Indeed, especially with the deep learning models, the resulting loss function to minimize is simply not tractable. This means the possibility of solving the optimization problem analytically is out of the question.

\subsection{Evaluating a machine learning algorithm}

The true intent behind using machine learning models is often not to minimize/maximize the actual loss used in training. Indeed, this mathematical formulation of the true goal is a surrogate loss that can be easily optimized because most of the time, the real measure/objective is not.

A key example of this is the classification task. The actual objective of doing classification is to minimize the number of errors a model makes on a new data sample it never trained one, i.e. generalization error. However, for the gradient optimization algorithms to work, the mathematical formulation of the objective must be continuous and differentiable, where the generalization error may not always be.

\subsection{Gradient optimization}

Given the complex nature of the loss functions used throughout , rather than finding analytically the solution, an iterative method can be used. If analytically solvable, the optimization procedure would have called for finding where the $\nabla_\theta J(\theta) = 0$ , where the gradients of the objective with regards to the parameters $\theta$ are zero. An intuitive alternative to circumvent the need to solve this is to consider an iterative solution, called gradient descent/ascent \cite{cauchy1847methode}.

\subsubsection{Stochastic Gradient Descent}

The gradient descent procedure can be conceptually understood as hiking down a mountain, where we do not know the actual path down. An intuitive way to go about reaching the bottom of a mountain would be at each step, to look for the angle of the mountain and take a step in the direction that goes downhill. If the mountain is \textit{convex} such that there are no valleys that restrict access to the bottom of it, this approach is guaranteed to allow you to reach the bottom of the mountain. 

This is in fact exactly minimizing the empirical risk through gradient descent. Iteratively, at each step, the direction that goes downhill will be computed and then a step will be made in that direction. The direction of the \textit{mountain} will be given by the gradient $\nabla_\theta J(\theta)$. To mitigate risks of getting stuck in local minima during the optimization, rather than computing the gradient based on the full training state, noisy estimates of the gradients can be computed by randomly selecting a subset of the data. This approach is commonly called minibatch stochastic gradient descent, where a minibatch of size $m$ represents a randomly selected subset of the data. The gradient estimate can therefore be defined as,

\begin{equation}
\hat{\nabla}_\theta J(\theta) = \dfrac{1}{m}\sum \limits_{i = 1}^{m} \nabla_\theta L(f_\theta(x^{(i)}), y^{(i)})
\end{equation}

In the most extreme case where $m=1$, the gradient can be estimated with a single sample from the data. A stripped-down version of minibatch SGD and its training algorithm is described in algorithm \ref{alg:stochastic-gradient-descent}.

\begin{algorithm}
    \caption{Minibatch stochastic gradient descent}
    \label{alg:stochastic-gradient-descent}
    \begin{algorithmic}[1] % The number tells where the line numbering should start
    
    \State Given $D_{train}=\{(x^{(1)}, y^{(1)}), \dots, (x^{(N)}, y^{(N)})\}$ the training dataset
    \State Given $f_\theta$ is a continuous and differentiable function
    \State Initialize $\eta$ as the step size
    \State Initialize $m$ as the size of the minibatch
    \While{Not converged}
    	\State Randomly select $m$ data samples from $D_{train}$
        \State $\theta \leftarrow \theta - \eta \dfrac{1}{m} \sum \limits_{i=1}^m \nabla_\theta L(f_\theta(x^{(i)}), y^{(i)})$
        
    \EndWhile
    \end{algorithmic}
\end{algorithm}

One key aspect of SGD is that the gradients must be evaluated with the same value of $\theta$ for the full minibatch. This is often one of the bottlenecks in the speed of training. As the minibatch size increases, the \textit{accuracy} of the gradient estimate increases, but so is the time to compute it. Although GPU implementations have been able to parallelize some of these computations to speed up training, other approaches have been proposed to further alleviate the time constraint.

\subsubsection{Synchronous and asynchronous SGD}

Distributed SGD refers to the widely used approach of data parallelism with large neural network models/datasets. The main concept is to split the data among different computation nodes and allow them to be connected to a central controller. Each node always has the same version of the model, and each is responsible to provide to the central controller the changes to the parameters for its local data share. To create a single parameter update of the \textit{global model}, each of the node computes the gradients on their local minibatch of data which comes from their local dataset partition. The gradients are then sent to the central controller, where the gradients from all compute nodes are agglomerated to form the global gradient estimator. The global gradient estimator is then sent back to all compute nodes, so they can all update their local version of the model in the same manner. This ensures each local model has the same version of the parameters as the others.

This method scales almost linearly with the number of computing nodes (up to a certain amount of computing nodes \cite{goyal2017accurate}). The computational advantage of this approach comes from the parallelism of the computation over multiple data partitions. The theoretical gains for a dataset split into $N$ partitions is $\frac{1}{N}$, because it is assumed each compute node can train on their local share in parallel. Furthermore, asynchronous versions \cite{bengio2003neural} of distributed SGD have been developed to alleviate some of the computational needs of the approach. However, synchronous methods have shown some limitations to scalability, leading to alternative methods such as co-distillation \cite{anil2018co-distillation}.

Throughout this document, this distributed SGD approach is considered as an approach leveraging communication between models because of the use of the central controller. In fact, this is considered to be the most efficient in terms of computation gains/speedup. The main aspect of distributed SGD that causes problem is the high bandwidth requirement that derives from sharing the \textit{raw} gradients. Given the size of the models used under this approach are usually large, different computation nodes are physically separated, therefore requiring a top-of-the-line network between compute nodes \textbf{and} the central controller.

\section{Neural networks and deep learning}

With machine learning being a subcategory of artificial intelligence, neural networks and deep learning are a subgroup of techniques and training algorithms of machine learning. They derive their name by the structural inspiration to the brain topology and how these algorithms leverage the \textit{stacking} of different layers of units.

All of the recent deep learning architecture base themselves on the linear combination of multidimensional inputs and parameters. Given an input $\textbf{x} = \{x_1, x_2, \dots, x_d\}$ and parameters $\textbf{w} \in \mathbb{R}^d, b \in \mathbb{R}$, the most basic of models to approximate a function of $\textbf{x}$, the linear combination, can be written as,

\begin{equation}
f(x) = \sum\limits_{i=1}^d w_i \times x_i + b
\label{eq:linear-combination}
\end{equation}

This model alone could be applied to previously mentioned tasks like classification or regression. In any of those two cases, $f(x)$ would serve as an approximation of the label or value $y$ and it could be trained by minimizing the empirical risk using minibatch gradient descent.

\subsection{Perceptron}

The ancestor of neural networks is the Perceptron \cite{mcculloch1943perceptron} which is a clever take on the linear combination in order to apply it to a binary classification class. Consider a dataset for a two group classification task with $D_{train} = \{(\textbf{x}^{(1)}, y^{(1)}), \dots, (\textbf{x}^{(N)}, y^{(N)})\}$, where $x \in \mathbb{R}^d , y \in \{ -1, 1\}$ in addition to parameters $\textbf{w} \in \mathbb{R}^d, b \in \mathbb{R}$. To simplify the notation, all parameters of a model can be grouped into a single variable $\theta = \{ \textbf{w}, b\}$.
\newpage
The perceptron defines the predicted class $f(x)$ as the following,

\begin{equation}
f(x) =
\begin{cases}
	1, & \text{if } \sum\limits_{i=1}^d w_i \times x_i + b > 0 \\
    -1, & \text{otherwise}
\end{cases}
\end{equation}
To minimize the empirical risk, the Perceptron used a creative loss function that allows it to employ gradient optimization algorithms. Indeed, as previously discussed, directly minimizing the number of errors on the training set is not feasible with SGD. However, the Perceptron proposes to slightly modify the error count in order to allow for the gradient computation. Denoting $h(x) = \sum\limits_{i=1}^d w_i \times x_i + b$, it can be expressed in the following manner,

\begin{equation}
L(f_\theta(x^{(i)}), y^{(i)}) = \mathbf{1}_{\{ y^{(i)} \times f_\theta(x^{(i)}) \le 0 \}}(- y^{(i)} \times h(x^{(i)}) )
\end{equation}
The loss function is therefore valued at 0 unless there is an error in the prediction, in which case it is equal to the linear combination of the parameters and the input $\textbf{x}$. Without getting into the details, this loss function can be plugged in the previously described optimization algorithms to iteratively train $\theta$.

\subsection{Non-Linearity and activation functions}

The Perceptron can be decomposed into two functions quite intuitively. Firstly, a linear combination of the parameters, or weights and the input that gives $h(x) = \sum\limits_{i=1}^d w_i \times x_i + b$, and the step function that gives the final output $f(x) = \text{sign} (h(x))$.

\begin{figure}[b]
  \centering
  \captionsetup{width=.8\textwidth}   
  \includegraphics[width=.8\textwidth]{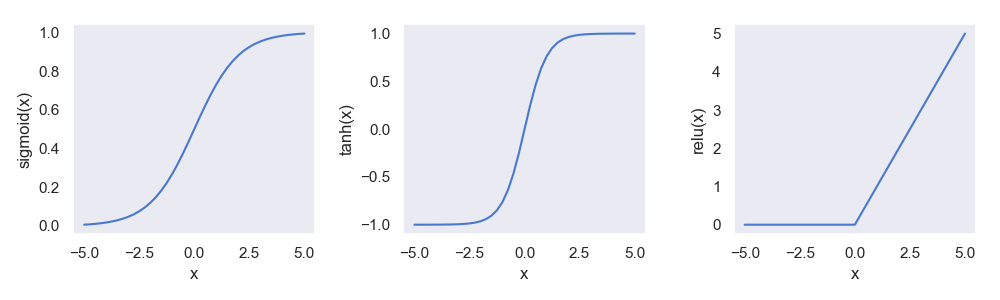}

  \caption{Popular non-linearity functions used in neural networks. (Left) Sigmoid, (Middle) Tanh and (Right) ReLU.}
  \label{fig:non-linearity}
\end{figure}

Similarly, any function could be applied to $h$ in order to get an output. These functions are called non-linearity or activation functions and are a key concept of allowing neural networks to become high capacity models. Without the use of a non-linearity, the model can only represent linear relationships. Although any function could be used as a non-linearity, given the use of gradient optimization, it is required that it be differentiable almost everywhere. 

The ReLU \cite{nair2010rectified}, the sigmoid and tanh functions are the most common activation functions. They serve different purposes, with the ReLU often used as a hidden layer activation, while the sigmoid can be used to provide a value between $0$ and $1$, like a probability. The tanh function used to be a popular hidden layer activation function, but it is mostly used now to force values between $-1$ and $1$. Theses non-linearities are plotted in Figure \ref{fig:non-linearity}, and their functions are as follows,

\begin{itemize}
\item \textbf{ReLU}: $f(x) = \max(0, x)$
\item \textbf{Sigmoid}: $f(x) = \dfrac{1}{1+e^{-x}}$
\item \textbf{Tanh}: $f(x) = \dfrac{e^{x} - e^{-x}}{e^{x} + e^{-x}}$
\end{itemize}

\subsection{Feedforward Neural Network and Multi-Layer Perceptron}

The linear combination described in equation \ref{eq:linear-combination} allows one to combine an array of inputs $\textbf{x}$ and parameters $\textbf{w}$ and b into a single scalar. A natural way of expanding this is by considering multiple values of parameters $\textbf{w}$ and b. If indeed, instead of having a single linear combination with a single set of $\textbf{w}$ and $b$, there were $k$ linear combinations with each their corresponding parameters, $f(x)$ could therefore become multidimensional. In addition, much like the Perceptron, a non-linearity function could be applied to each of these.

For simplicity, a linear combination of the parameters and an array of inputs, in combination with a non-linearity, can be called a node or a unit. The analogy of the neural network then comes from the visual representation where a single node is therefore \textit{connected} to the input array.

The combination of the $k$ units or nodes is referred to as a layer, and the $k$ scalar values \textit{generated} from the computations are called the outputs of the layer. In a way, this layer is composed of $k$ Perceptrons, allowing the model to have an input dimension of $d$ and an output size of $k$. The $k$ outputs of the previously defined layer could be very well seen as an input themselves and in fact, this is the key concept behind neural networks. Indeed, layers can therefore be \textit{stacked} on top of each other, where the output of a previous layer is the input to the next. When considering the whole set of nodes, the last layer is referred to as the output layer, while all the others apart from the input are called hidden layers.

\begin{figure}[t]

\centering
\captionsetup{width=.8\textwidth}

\def\layersep{2.5cm}

\begin{tikzpicture}[shorten >=1pt,->,draw=black!50, node distance=\layersep]
    \tikzstyle{every pin edge}=[<-,shorten <=1pt]
    \tikzstyle{neuron}=[circle,fill=black!25,minimum size=17pt,inner sep=0pt]
    \tikzstyle{input neuron}=[neuron];
    \tikzstyle{output neuron}=[neuron];
    \tikzstyle{hidden neuron}=[neuron];
    \tikzstyle{annot} = [text width=4em, text centered]

    % Draw the input layer nodes
    \foreach \name / \y in {1,...,3}
    % This is the same as writing \foreach \name / \y in {1/1,2/2,3/3}
        \node[input neuron, pin=left:$x_\y$] (I-\name) at (0,-\y-1) {};

    % Draw the hidden layer nodes
    \foreach \name / \y in {1,...,6}
        \path[yshift=0.5cm]
            node[hidden neuron] (H-\name) at (\layersep,-\y cm) {};

    % Draw the output layer node
    \node[output neuron,pin={[pin edge={->}]right:Output}, right of=H-3] (O) at (2.5, -3){};

    % Connect every node in the input layer with every node in the
    % hidden layer.
    \foreach \source in {1,...,3}
        \foreach \dest in {1,...,6}
            \path (I-\source) edge (H-\dest);

    % Connect every node in the hidden layer with the output layer
    \foreach \source in {1,...,6}
        \path (H-\source) edge (O);

    % Annotate the layers
    \node[annot,above of=H-1, node distance=1cm] (hl) {Hidden layer};
    \node[annot,left of=hl] {Input layer};
    \node[annot,right of=hl] {Output layer};
\end{tikzpicture}

    \caption{Visual representation of a multi-layer perceptron with a single hidden layer with 6 units, input $\textbf{x} \in \mathbb{R}^3$ and a single output.}
    \label{fig:mlp}

\end{figure}
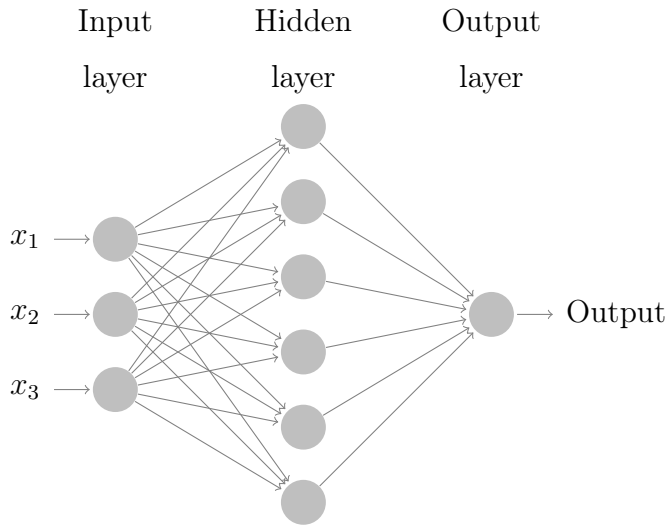

Names often given to these networks are the feedforward neural network and the Multi-Layer Perceptron (MLP), as per their structural design. A visual representation of a single hidden layer MLP can be seen in Figure \ref{fig:mlp}.

Given enough memory to store all the parameters, these models can be built of arbitrarily size, where both the number of units and the number of layers can be controlled. Having multiple layers and numerous units allows for the output of the network to be a highly complex function of the input. The MLP has an interesting property: provided with an infinite number of hidden units, it is a universal function approximator \cite{hornik1991mlp-approximation}. In other words, it can have enough complexity through the combinations of the input that any continuous function can be reproduced.

\subsection{Convolutional Neural Networks}

Convolutional neural networks \cite{lecun1990convolution} (CNN or ConvNet) are a special type of neural networks especially designed to handle data such as images. CNNs have been shown to be applicable not only to images but also to text \cite{johnson2014cnn-text} and even audio signals \cite{van2016wavenet}. These models are what popularized deep learning by their highly optimized implementation and impressive performance on difficult image classification tasks such as ImageNet \cite{deng2009imagenet}.

The key concept behind CNNs is the parameter sharing aspect that allows for the same computations to be made at different places of a layer. In comparison, an MLP layer needs to have specific computations for all of the layer's input dimensions. This is very useful when applied to images because it allows the model to \textit{detect} the same shapes and patterns throughout the image. These models have the property of being shift and space equivariant, which simply means that through the convolution operation of a layer, a shift in the input will result in the same shift in the output.

It is important, however, to point out that this type of network can very well be, and is most often, mixed with the MLP. In fact, the different types of networks are usually handled as layers, where it doesn't matter how a given output was obtained, as long as it can be considered as an input to another layer. Multiple convolutional layers are often used to extract features from images, only to be combined higher in the architecture with an MLP.

\subsection{Recurrent Neural Networks}

Another type of neural network called Recurrent Neural Networks (RNN) \cite{rumelhart1986rnn} are used to handle sequential data such as time series or video. To handle the input changing over time, RNNs have the characteristic of sharing parameters \textit{through time}. Indeed, rather than having a separate set of parameters for each time step, it uses the same parameters for each time step.

Given an input $\textbf{x} = \{x_1, \dots, x_t, \dots, x_T\}$, where $x_t \in \mathbb{R}^d$ and parameters $W_x \in \mathbb{R}^{d \times m}, W_h \in \mathbb{R}^{m \times m}, \textbf{b} \in \mathbb{R}^m$, it can be written as a recurrence, in the simplest case as,

\begin{equation}
\textbf{h}_t = f(\textbf{x}_t, \textbf{h}_{t-1}) = W_x \bullet \textbf{x}_t + W_h \bullet \textbf{h}_{t-1} + \textbf{b} \\
\end{equation}

With a hidden state $\textbf{h}_t$ computed for each time step, it can be used as an input to another layer to generate an output. This is an example of the concept of layers that is key to understand how neural networks are built. Furthermore, it can be easy to imagine a wide array of combinations of inputs and outputs, especially given the added dimension of time that is often associated with RNNs. The most widely used architectures for RNNs are the long short-term memory \cite{hochreiter1997lstm} (LSTM) and gated recurrent units \cite{cho2014gru} (GRU). Both of these models have shown to allow for longer dependencies between time steps to emerge by limiting the effects of the exploding and vanishing gradients, which have been described to cause problems to the optimization procedure \cite{bengio1994learning}.

\subsection{Unsupervised learning models}

In this section, the most popular deep unsupervised learning models are detailed which were also used in the subsequent chapters of this document.

\subsubsection{Variational Auto-Encoder}

The Variational Auto-Encoder \cite{kingma2013vae} (VAE) is a model of the encoder-decoder type that allows to both encode an input into features, but also generate samples. An encoder network maps from the input space of $x$ to $z$, the feature space, which can be of arbitrary size. The true distribution of $z$ given an input $x$ can be denoted $p(z|x)$ and the encoder's model of that distribution will be defined as $q(z|x)$. It turns out by its training objective, $q(z|x)$ will be trained to become closer to its true posterior distribution $p(z|x)$. This in turn makes it possible to sample $z \sim q(z|x)$. Another network, the decoder, maps the sampled $z$ back into the input space as a generator. The output distribution of the reconstructed inputs can be denoted $p(x | z)$.

The key and the beauty to making this whole model work is the training objective which is called the variational lower bound or evidence lower bound (ELBO). This bound to the log-likelihood of the underlying distribution of the data $p(x)$, can be maximized to create meaningful features and a good sample generator. Using the above mentioned distributions, it can be defined as,

\begin{equation}
\begin{split}
\mathcal{L}(q) &= \mathbb{E}_{q(z|x)}\big[\log p(x, z)\big] + H\big(q(z|x)\big) \\
& = \mathbb{E}_{q(z|x)}\big[\log p(x | z)\big] - D_{KL} \big[ q(z|x) || p(z) \big]  \\
& \le \log p(x)
\end{split}
\end{equation}

Practically speaking, the left most objective, $\mathbb{E}_{q(z|x)}\big[\log p(x | z)\big]$ is attributed to the decoder and is considered to be the reconstruction or prediction error. Indeed, it is trained to increase its \textit{ability}, given a sampled $z \sim q(z|x)$, to predict $x$. The rightmost part $D_{KL} \big[ q(z|x) || p(z) \big]$, the Kullback-Leibler (KL) divergence, will make the features generated by the encoder closer to the prior distribution $p(z)$ which is defined as a Gaussian distribution. This approach has the advantage, depending on the task on hand, to encode an input into a distribution, rather than a simple deterministic mapping.

\subsubsection{Generative Adversarial Networks}

GANs, short for generative adversarial networks \cite{goodfellow2014generative} revolutionized the world of generative models. Although numerous variants of the \textit{vanilla} GAN have been proposed, this brief section details the original version.

The main concept at the center of this type of model is the competition between two components, the generator and the discriminator. Firstly, the generator with its parameters $\theta_G$, is a function of a noise vector $z \in \mathbb{R}^m$, and outputs directly fake samples, i.e.  $G_{\theta_G}(z) = x'$. The discriminator, with its own set of parameters $\theta_D$ takes as input either the original data $x$ or the fake sample $x'$, and gives a single scalar score between $0$ and $1$ using a sigmoid non-linearity on the output layer. The discriminator network is noted as $D_{\theta_D}$. Conceptually, this score represents the confidence of the discriminator that the provided input (whether real or fake) is a real sample.

The discriminator is trained in order to both increase the score of true data decreasing the score of the fake data. In a way, it is learning to distinguish between the true distribution of the data $p(x)$, and the fake distribution of data $q(x |z)$. The adversarial aspect of the model derives from the way the generator is trained. Indeed, its objective is to compete with the discriminator and generate samples that would be considered as true. In other words, it is trained to fool the discriminator into thinking its samples are real.

Given samples $x \sim p(x)$ and $x' \sim q(x|z)$ with $z \sim q(z)$, from the true data distribution and generated samples, respectively, the global objective can be written as,

\begin{equation}
\mathbb{E}_{p(x)}\big[\log D_{\theta_D}(x)\big] + \mathbb{E}_{q(x|z)}\big[\log (1- D_{\theta_D}(x'))\big]
\end{equation}

Relating back to the adversarial aspect, the discriminator is trained to maximize this full objective, while the generator will be trained to minimize it.

\subsubsection{Bidirectional Generative Adversarial Networks}

One disadvantage of the GAN is that it doesn't provide for a \textit{compressed} representation of the data. Unlike the VAE, it \textit{only} build a model that allows to generate some samples. Two similar models have been designed to leverage the adversarial aspect of GANs in order to generate features out of input data. Although Adversarially Learning Inference \cite{dumoulin2016adversarially} is similar to Bidirectional Generative Adversarial Networks \cite{donahue2016adversarial} (BiGAN), the latter are slightly more straightforward to explain and was used in some experiments detailed in this document.

The key difference between GAN and BiGAN is the added encoder $E_{\theta_E}(x)$ that is used to map true data samples to the feature space. Previously, the discriminator was only fed either the true image, or the generated samples from the generator. In BiGAN, the noise vector and the encoded features are paired up along with their corresponding generated and true data samples, respectively. Much like the traditional GAN framework, the pairs are then considered as the fake and true data and fed to the discriminator.

Training of the generator and discriminator is the same as in the original GAN formulation, while the encoder is also trained to minimize the objective. The authors of BiGAN further argue that by using the proposed objective, the encoder learns to invert the generator.

\section{Capacity and regularization}
\subsection{Controlling the capacity}

The capacity of a model refers to its ability to represent a large space, or family, of functions. Although abstract, it can conceptually be understood as a measure of how \textit{flexible} a model is. For example, a model such as a deep learning network with many parameters has high capacity and is therefore known to be able to represent highly complex functions, while a linear model has very low capacity since it can only represent linear relationships.

Capacity can often be controlled by the choice of machine learning algorithms, but even within the configuration of a particular algorithm. It could seem intuitive to always aim for the highest possible capacity when tackling a machine learning task, however there is an important caveat and it relates to the objective being optimized. Given the training procedures minimizes the empirical risk rather than the \textit{true evaluation} objective, and although increased capacity should allow to reach the minimal training loss, it might not be ideal in terms of generalization error.

\subsection{Ensemble learning}

Ensemble learning is a machine learning method that allows for combining a set of models into a single one. A common approach, called model averaging, consists of fully training variants of the same machine learning model on the same data and then combining them when making a prediction. This has the effect of leveraging the diversity in the different solutions that are proposed by the set of machine learning models. An early version of this approach, called bootstrap aggregating (or bagging) \cite{breiman1996bagging}, proposed to train models on different subsets of the training data. 

This is another type of algorithm that is considered to have a communication protocol. At the time of inference or deployment of the ensemble of models, a communication occurs since only a single prediction is made out of the ensemble. All the models which are part of the ensemble therefore, in some way, do communicate in order to jointly make one prediction.

\subsection{Regularization techniques}

Regularization consists of limiting the capacity of the model such that optimizing the training objective does not make performance on the \textit{evaluation} objective worst.

\subsubsection{Weight-decay}

A widely used regularization approach is to impose some constraints on the different parameter weights of neural networks. This is done by adding an additional objective to the empirical risk $J(\theta)$. The additional objective to minimize can be the square norm of the weight vectors. Since minimizing the norm of the vector goes \textit{against} the main objective, it is considered to be restricting the model as it \textit{pushes} all of the parameters values towards 0. Furthermore, the importance of this additional objective is controlled by a hyper-parameter $\lambda$, where a larger value indicates a greater importance in comparison to the original empirical risk.

Indeed, by considering the L1 or the L2 norm of the weight vectors, the global objective can become any of the two objectives described below. Additionally, in some cases such as elastic net regularization  \cite{zou2005regularization} both the L1 and L2 norms may be combined to ensure proper regularization of the network's weights.

\begin{align}
J_1(\theta) &= J(\theta) + \lambda_1 \sum\limits_i |\theta_i| \\
J_2(\theta) &= J(\theta) + \lambda_2 \sum\limits_i \theta_i^2
\label{eq:weight-decay}
\end{align}

\subsubsection{Dropout}

The dropout \cite{srivastava2014dropout} regularization approach is a very interesting and simple approach that provides great regularization power. It consists of simply injecting noise to a layer by sampling a binary mask on the input and hidden layers. For example, given $x = \{ {x_1, x_2, \dots, x_d} \}$, a binary mask of size $d$ would be sampled from a Bernouilli distribution with probability $p$. The amount of noise is controlled by the probability $p$, a hyper-parameter.

During training, the mask is sampled for each $x$ that \textit{goes through} a given layer of a network and the corresponding dimensions of $x$ that match the noisy mask are turned off. During evaluation, the probability is set to 0 and inference is made on the full observations for each layer. By sampling different masks at each layer, it has the effect of limiting the number of \textit{pathways} in the neural network. Furthermore, this approach has been compared to training an ensemble of models \cite{baldi2013dropout-ensemble} at a much less expensive resources cost.

\subsubsection{Early stopping}

A widely used method of monitoring the progress of the training procedure of a network is to consider how it performs on the actual objective. For example, by monitoring the generalization error of a network used for classification during training and stopping it when the model no longer improves is a way of regularizing it. Indeed, by limiting the amount of gradient optimization steps it makes, the number of functions that can be represented by the network is limited.

\chapter{Exchanging Outputs Between Models\label{chpt:exchg-outs}}
% Exchanging Outputs Between Models

% Abstract / little summary of what is done
Most implementations of large scale neural networks are trained using either \textit{synchronous} \cite{dean2012large, zinkevich2010parallelized} or \textit{asynchronous} \cite{bengio2003neural} gradient optimization procedures. Although both of these approaches have some great benefits and can achieve impressive results leveraging data parallelism, they require a great deal of both computing and networking resources to actually achieve their theoretical optimal speedup. In this work, we propose to explore how removing one of the main characteristics of these optimization procedures may affect performance. Mainly, this work studies how decentralizing the computation onto multiple compute nodes and allowing them to communicate between each other rather than to a central controller, may reach similar performance. Part of the study includes what information can be communicated between the compute nodes, which includes low bandwidth outputs and hidden layers activation.

% Intro / Motivation
\section{Introduction}

Previous work on allowing a model's output to be used as training labels for another one, known as distillation \cite{hinton2015distilling}, have paved the way to leveraging a network's outputs to accelerate training in a teacher/student setup. More recently, \cite{anil2018co-distillation} showed distillation could be applied online during training on two large models in order to circumvent some of the scaling issues of distributed synchronized SGD. Building on this, this work explores how this expands to a full network of computation nodes that exchange outputs at different depths of their respective architectures.

The current shift towards using \textit{synchronous} and \textit{asynchronous} gradient optimization as deep learning models and datasets grow bigger entails high bandwidth requirements to trade off the large local computations done at each node. In some cases where highly optimized infrastructure and large computational power are available, distributed-SGD can drastically reduce the time needed to train large deep learning models \cite{goyal2017accurate}. There are therefore large gains of bandwidth to be made by decentralizing some of the computation.

Reconsidering a decentralized version of the distributed-SGD further allows for a reconsideration of future computational needs and possibilities. Using a decentralized computing network could allow for increased stability and failure resistance as the whole progress of training would not be dependent on the success of all the compute nodes. Such an approach could also lower the cost of training large deep learning models through the lack of a requirement of highly optimized and efficient computing centers.

In a way, all of these potential advantages explored here could be extended to form an \textit{internet of computing}. One could indeed imagine a large network of low powered and low bandwidth computing nodes exchanging information between them without the need to centralize the communication, all working on the same objective.

An inspiration for studying the effects of decentralizing the communication between models comes from society and the way humans go about communicating with each other. There is a parallel to be made with the way all humans exchange with each other directly when the intent is to learn something. In particular, when the wide range of knowledge/data available to the society to be known is considered, an analogy can be drawn regarding specialization of the different \textit{members} of society for different subsets. When called upon to solve a problem, the society unites to propose a solution. This work touches on parts of this analogy throughout the way it trains, communicates and predicts.

As part of the decentralized and communicating network of computing nodes, different components and parameters will be explored. In particular, which nodes can communicate, what they communicate and how often they communicate with each other will be part of the tested configurations of the computing group. Both supervised and unsupervised learning tasks are used in order to analyze the effectiveness of the proposed approach. An overview of the experimental setup is, however, required to show the different components at play.

\section{Method}

\subsection{Simulated nodes}

In order to alleviate the engineering and computational needs often associated with using a \textit{physical} network of computing nodes, all experiments were done with models in a \textit{simulated network of computing nodes} on the same physical machine. What is meant by \textit{simulated} is that the computing nodes were not truly setup as a network of computing nodes but rather all stored on the same physical machine. Using academic resources such as large clusters of GPUs for relatively high computational needs is possible, but to require the exclusivity and availability of more than 50-100 GPUs across multiple machines was simply not feasible. Furthermore, having all the \textit{simulated nodes} effectively on the same machine allowed to simplify the implementation of the communication protocol between nodes.

Ultimately, the main limitation to both training speed and memory usage of the implementation was the number of nodes in the network. The approach chosen to be able to scale up efficiently with the number of nodes was to have the training loops for the compute nodes to be sequentially evaluated. Some significant operations such as the data management and the communication protocol were implemented in parallel throughout all the compute nodes in order to be able to shave off the sequential overhead. Given the appropriate resources, all experiments could be extended relatively easily to a full network of parallel computing nodes.

\subsection{Data and experimental setup}

\subsubsection{Dataset}

To test the effect of decentralized communication and to accommodate the practicality of experiments, the dataset used was the CIFAR-10 dataset. This dataset consists of 50,000 colored images of size 32x32 used for training as well as a distinct set of 10,000 images of the same size used as the test set. Each of the images represents 1 of the 10 different classes uniformly split through both the training and test sets.

To reflect the decentralization of the data between computing nodes, the entire training set was randomly split among the different computing nodes in the network. In other words, given a network consisting of 25 nodes, each of the node has their own distinct, and exclusive, 25th of the training data. Furthermore, the distribution of data among the computing nodes was done independently of the classes. The fractional training data that each computing node has access to is considered, and referred to, as the \textit{local} training data.

In the actual implementation, given the simulated network of nodes, the dataset was centralized to the physical machine used to hold all the models as it provided easier access and simplified experimental design. In no way was any of the \textit{effective nodes} using the partition of another node's training data. Much like the extension to physically distinct computing nodes, the extension of the dataset setup could be done to be truly decentralized.

\subsubsection{Model architecture}

With the dataset this work focuses on, the logical choice of a model was to use Convolutional Neural Networks (CNNs). These models are widely used in the deep learning community when the input are images, given their optimized GPU implementation but in particular their structural characteristics. Such characteristics include parameter sharing and their ability to be invariant to slight input transformations such as translations.

The size of the model and the details of the architecture of the CNNs used throughout the experiences are not essential to the understanding of this work. For completeness, the models were traditional CNNs with no pooling layers but rather strided convolutions. For work in the unsupervised learning setting, if a decoder was necessary, the same structure as the encoder was used. In general, the structural recommendations from the DCGAN \cite{radford2015dcgan} architecture were followed.

\subsubsection{Communication and network of nodes}

In order to design the communication exchanges between compute nodes, one key aspect to be thought of is how each of these computing nodes will communicate throughout training. The first thing that needs to be considered is which node can communicate with each other. There is, of course, a combinatorial way of designing sets of nodes that can communicate with each other. To simplify this, let's consider the analogy to human communication at both extremes. On the one hand, there is how each of us communicates with a small set of relatives, but there is also the opposite where we attend classes or conferences where the same information is distributed to a much wider array of individuals.

Implicit to these is the distinction between the \textit{broadcasting} and the \textit{consumption} of information. One could therefore be broadcasting to a large number of nodes, e.g. giving a talk at a conference, or to a small number of nodes, e.g. speaking to close relatives. Regarding consumption, attending a conference would allow for consumption from a large array of different sources, while being exposed to close relatives would restrict that number.

To address this in the implementation, the communication between nodes was designed such that a given node can broadcast to its $p$ neighbours, $p$ being controlled as a hyper-parameter. As for the consumption, it is controlled implicitly by the dynamics of the network resulting from the hyper-parameter $p$. For example, in a 25-node network if $p$ is set to 24, this means that a given node can broadcast to all other nodes, and every node can broadcast to all other nodes. The set of nodes to which a given node can broadcast is considered to be its \textit{neighbours}. In a network of nodes with a much more restricted communication, e.g. consider a 100-node network with $p = 5$, the constraint imposed in the implementation is that for a node, its 5 neighbours must be \textit{adjacent}. An illustrative way of understanding this is by considering all nodes in a circle, with the neighbours being selected as the closest nodes.

For simplicity, the communication pattern between nodes is considered fixed and $p$ the same for all nodes. A possible extension of this work consists of using more complex sets of connections, e.g. each node has a random set of neighbours, fixed or changing.

In order to further control the communication between nodes, frequencies of broadcasting and consumption were added as hyper-parameters. Simply put, for a given minibatch, the broadcasting frequency can be seen as the likelihood to broadcast to it's neighbours. As for the consumption frequency, it can be seen as the likelihood of consuming data sent from the other nodes rather than from its \textit{local} training data. In addition, if a node is training on data sent from another node, it is also exposed to the possibility of being broadcasted itself, i.e. broadcasted data can be broadcasted to other nodes.

To implement data consumption from other nodes, each has their \textit{local} training data in addition to a consumption queue where all the data sent from other nodes is added to. This consumption queue is analogous to an \textit{email inbox} that will be receiving and storing all data locally where it will be read from. All the components impacting the communication of a single computation node can be effectively summarized in Algorithm \ref{alg:consume-broadcast}.

\begin{algorithm}
    \caption{Consumption and Broadcasting training pseudo-code (for one node)}
    \label{alg:consume-broadcast}
    \begin{algorithmic}[1] % The number tells where the line numbering should start
    
    \State Initialize $q_{local}, q_{external}$ as local training data queue and empty external data queue
    \State Initialize $p_c, p_b$ as consumption and broadcasting probabilities
    \State Initialize communication channel to $q_{external}$ of the neighbours
    \While{Training}
        \State $consume \sim \text{Bernouilli}(p_c)$
        \State $broadcast \sim \text{Bernouilli}(p_b)$
        \If{$consume$}
            \State $data \gets $ pop $q_{external}$
            \State Do consumed data training objective step
        \Else
            \State $data \gets $ pop $q_{local}$
            \State Do local training objective step
        \EndIf
        
        \If{$broadcast$}
            \For{all neighbours}
                \State Put $data$ in their $q_{external}$
            \EndFor
        \EndIf
        
    \EndWhile
    \end{algorithmic}
\end{algorithm}

\subsubsection{Collective decision making \label{sec:collective_decision}}

As communication during training was detailed in the previous section, another key aspect is to consider how each of the computing nodes will communicate at test time. In this line of work, all the nodes have a randomly initialized model and each has their own subset of the data, but the focus remains on combining the knowledge from each of the nodes and consider their \textit{total knowledge} as a group. In addition to being aligned with the human culture analogy previously described, traditional and centralized approaches leveraging data parallelism usually aim to train a single model to make a single prediction at test/inference time.

Traditional ensemble methods such as bagging \cite{breiman1996bagging} that make accessible the full dataset to each of the models combine predictions by averaging results or if applicable, by using a voting scheme. Under the studied framework, the preferred approach was to employ a form of weighted model averaging. The weighting is done at the level of the output probabilities. For example, in the supervised setting, all nodes are presented with the data to make a prediction on and they \textbf{all} provide their distribution over the possible answers, mainly, the different classes along with their corresponding probabilities. The distributions are then gathered for all nodes, the entropy is then computed for each of the per-model distribution, and a single distribution is created by weighing each of them with their negative entropy. The entropy used is the Shannon entropy, leveraging its relationship to uncertainty as a confidence level for each node. The collective decision making algorithm is described in the algorithm \ref{alg:collective-decision}.

\begin{algorithm}
    \caption{Collective decision making pseudo-code}
    \label{alg:collective-decision}
    \begin{algorithmic}[1] % The number tells where the line numbering should start
    \State Input $x$ is received by every compute node
    
    \State Each of the nodes compute their probability distributions $y^i=[y^i_1, y^i_2, \dots, y^i_C]$
    \State Collect all $y^i$'s and initialize sum of negative entropy $s= 0$
    \For{All nodes and their corresponding $y^i$}
    	\State Compute entropy as $h^i = - \sum_k y^i_k \log y^i_k$
        \State Increment total sum of negative entropy $s = s - h^i$
    \EndFor
    \For{All nodes}
    	\State Compute normalized weight based on negative entropy as $w^i = -h^i / s$
    \EndFor
    \State Compute single weighted distribution as $\textbf{y} = \sum_i w^i \times y^i$
    \State Make a single class prediction as $\operatorname*{argmax} \textbf{y}$
    \end{algorithmic}
\end{algorithm}

The prediction is therefore made out of the entropy weighted average distribution between nodes. This work makes an important assumption at test time; that all nodes are \textit{reachable} in order to make a prediction. Although not explored here, future work on collective decision making at a much larger scale should consider applying the same approach but from collecting predictions from only a subset of the nodes rather than all the network nodes. If this framework were to be extended to a very large number of computing nodes such as the \textit{internet of computing}, requesting an answer from all nodes would simply not be feasible.

\subsection{Supervised Setting}

\subsubsection{Training objective and evaluation}

Regarding a single node along with its local training data, the supervised learning procedure and objective are standard. The objective for each node is to minimize the cross-entropy loss over all the local training data considering the corresponding label of each training image. If there is a communication channel between two nodes and depending on what information is communicated between the nodes, an additional training objective is considered; more on this in section \ref{sec:sup-learning_info-comm}.

As for evaluation, the network of nodes aims to have a low generalization error, much like all other traditional supervised learning tasks. In practice, the accuracy on the data left out of the training procedure is used as a measure of generalization performance. Considering that there is a full group of models rather than a single one to measure accuracy and that all the nodes need to be evaluated as a whole, the same data is used to evaluate all the nodes. The predictions on the data left out of the training procedure are made the same way as described in section \ref{sec:collective_decision}. A prediction is considered correct if the class associated with the highest probability is the correct one.

The measure of accuracy over training steps will be adjusted to reflect the acceleration potential of the approach. In other words, given that our implementation simulates a parallel system, some operations can be assumed to be potentially executed in parallel. Given an appropriate computation network, all training steps could be executed at the same time for all compute nodes.

\subsubsection{Information communicated between nodes \label{sec:sup-learning_info-comm}}

In order to avoid communicating directly the gradients between nodes or to a central system as with the \textit{synchronous} optimization algorithms, different outputs of each of the models in the network of compute nodes can be exchanged at different depths of the architecture. Given a classification task on hand, neural networks compute label predictions for each input as the highest level output. These label probabilities are a normalized version of what is commonly called \textit{logits}, or class label scores. The normalization of these logits is usually done with the softmax function. 

An intuitive thing to share between compute nodes would be the class label for a training sample. However, as previously described in distillation \cite{hinton2015distilling}, additional information about a model can be extracted in the logits and in turn, accelerate training of a secondary model if used as training targets, in particular when the temperature of the logits is raised. The modified logits can be normalized to create another predictive distribution of the labels and are further referred here as \textit{soft-labels}. The operation of normalizing the received logits into soft-labels is detailed in equation \ref{eq:raising-temp}.

Given the received logits $[v_1, v_2, \dots, v_C]$ from another model for each of the $C$ classes, and with temperature $\tau$, the soft-labels $[y_1, y_2, \dots, y_C]$ to be used as targets can be computed as, 
 
 \begin{equation}
y_i = \dfrac{\exp(v_i / \tau)}{\sum_j \exp (v_j / \tau) }
\label{eq:raising-temp}
\end{equation}

\begin{figure}[b]
    \centering
    \includegraphics[width=0.8\textwidth]{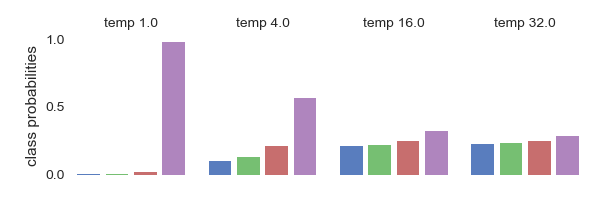}
     \captionsetup{width=.8\textwidth}
    \caption{Class probabilities resulting from normalizing the logits with different temperature. The logits or class scores used, i.e. 1.0, 2.0, 4.0, 8.0 for the 4 classes, are the same across the different temperatures above.}
    \label{fig:softmax_temp}
\end{figure}

As the temperature approaches zero, the soft-labels \textit{hardens} and becomes more like a \textit{one-hot} vector of the predicted class label. In contrast, as temperature approaches infinity, the soft-labels become uniform. See Figure \ref{fig:softmax_temp} for effect of varying the temperature and the resulting class probabilities. Conceptually, exchanging the soft-labels can be seen as exchanging what a node thinks the answer is, as opposed to sending directly the answer, i.e. the \textit{true label}.

Whether the information communicated is the \textit{soft} or true labels between nodes, both of them can be used with the same additional training objective. Indeed, for the node on the receiving end of such information, they can be seen just like regular training data with their corresponding labels. If the true label is exchanged, using the same training objective as for the local training data is straightforward. As for the soft-labels, the slight difference is simply to take into consideration the full probability distribution over the classes from the sender as being the target label, as shown in equation \ref{eq:cross-entr-full-distr}.

Using $y = [y_1, y_2, \dots, y_C]$ as the soft-labels or the true labels and $f_\theta(x)$ as the predictive probability distribution over the classes for a given input $x$, the cross-entropy loss originally defined in equation \ref{eq:cross-entropy} can be extended to consider the full probability distribution as,

\begin{equation}
L(f_\theta(x), y) = - \sum\limits_{j=1}^C y_j \log f_\theta(x)_j
\label{eq:cross-entr-full-distr}
\end{equation}

In addition to soft-labels and labels exchanged, exchanging high-level features between nodes was tested. Conceptually, the highest hidden layer before the label prediction output is an abstract representation of the image fed as input. This in itself makes it a good candidate for information to be shared between the models. Much like sending labels, whether soft or not, to another node, sending the features can be passed in the same way through the communication channel.

As for the training objective of another node's features, and given the features are trained under no constraint apart from the main supervised training objective, an appropriate loss function to use is the mean-squared error. Given features of an input $x$ from models $A$ and $B$ represented as $h^A = [h^A_1, h^A_2, \dots, h^A_m]$ and $h^B = [h^B_1, h^B_2, \dots, h^B_m]$, the MSE objective can be defined as,

\begin{equation}
L(h^A, h^B) = \sum\limits_{k=1}^m\frac{1}{2} (h^A_k - h^B_k ) ^2
\label{eq:full-mean-square-error}
\end{equation}

A receiving node is therefore trained to match its top-level representation to its neighbours’ by considering them as arrays of real values. From the group's point of view, exchanging top-level features consists of ensuring all the nodes of the network extract similar features from the same images.

\subsection{Unsupervised Setting}

A great way to test the effectiveness of the communication channel is to test it under an unsupervised learning setting. Previously in the supervised setting, it was still possible for a computing node to communicate its true label corresponding with the training data. In contrast, with the unsupervised task, no information apart from the training data themselves is assumed to be available. This then becomes a question of how a model can send good features to others in the computing network.

\subsubsection{Training objective and evaluation}

Much like in the supervised setting, all nodes train on their local data with their own respective objective. In this case, their objective is to extract meaningful features out of the images. Considering the intent is to communicate features, more on this in \ref{sec:unsup-learning_info-comm}, two \textit{families} of unsupervised models that allow for encoding of an input into features were explored:

\begin{enumerate}
\item Variational Auto-Encoder (VAE) \cite{kingma2013vae}
\item Bidirectional Generative Adversarial Networks (BiGAN) \cite{donahue2016bigan}
\end{enumerate}
Both of these allow for nodes to have their own local training objective and don't influence the general understanding of this work.

The VAE is comprised of an encoder and a decoder and is trained on the local data to both ensure the features extracted lead to a reconstruction of the input and also make the features themselves be similar to a Gaussian distribution. One key feature of the VAE is the fact the learned features are \textit{distributional}, as the encoder's output corresponds to Gaussian parameters $\mu$ and $\sigma$.

As for the BiGAN, it has three components. The generator is the same concept as in the traditional GAN \cite{goodfellow2014generative} framework, such that it generates fake samples out of sampled noise. The added component to BiGAN is an encoder that maps from the input space to a feature space. In addition, these features extracted from the \textit{real} input (not from the generator's output) are paired with the input before being fed to the discriminator. The pair of features and \textit{real} input is considered as real, while the sampled noise along with the corresponding generated samples are considered as fake for the discriminator. Much like in GAN, the discriminator is trained to distinguish between the real and fake, trying to create a bigger distance between the distributions of real and fake. On the other end, both generator and encoder are trained to fool the discriminator by feeding them their outputs. 

Evaluating unsupervised learning is in itself a field of research, but for this line of work, the focus was on leveraging the same collective decision making for evaluation as in supervised learning. In order to do so, a linear classifier was added on top of encoders of each of the computing node which was then trained on the full training data. Doing so allowed for evaluating purely how effective was the exchange of communication between nodes regarding the feature extraction process. Much like the supervised setting, the weighted predictive distribution along with the accuracy on data left out of the training data was also utilized.

Unlike in the supervised setting, here the focus is mostly on performance rather than actual acceleration.

\subsubsection{Information communicated between nodes} \label{sec:unsup-learning_info-comm}

In the unsupervised learning setting, there is no grounded information such as class labels to be exchanged between nodes. Therefore, in this setting,  the features are used as information to be communicated. For a compute node on the receiving end of the sender, this unsupervised task now becomes similar to a supervised task as it tries to reproduce another model's output. The hypothesis is that for a model, it is \textit{easier} to learn through a supervised objective than an unsupervised one.

There is however, more flexibility as to what the training objective can be for the features exchanged. In particular when a VAE is used, the features are distributional and characterized as a Gaussian distribution. As the goal is to have models with a similar representation for the same input, an objective such as the Kullback-Leibler (KL) divergence between two Gaussian distributions can be leveraged. Indeed, minimizing the KL divergence, between the sender's and the receiver's features can be seen as \textit{pulling} the latter features distribution towards the former's. As for the experiments with the BiGAN, much like in the supervised case, the mean-squared error was used as the training objective was used for the feature matching.

\section{Results}

To evaluate how successful this approach is at decentralizing the computation efficiently, different configurations of the network of nodes were explored. Almost all of the configurations and parameters of the network of nodes tested were in the supervised setting. The rationale behind this is that both unsupervised and supervised learning tasks in this setting share very similar aspects, in particular when the compute nodes exchanging features under the supervised task. It is expected that the most successful configuration under the supervised setting should transfer to the unsupervised task, especially given the dataset is the same and the models are similar in size and structure.

For this work, the most meaningful hyper-parameters of the network of nodes configuration are the content being communicated (logits, true labels, features), the temperature, if applicable, at which the sent logits are raised to, the frequency of broadcasting and consumption, the number of nodes as well as the number of neighbours each compute node has. Although not detailed here, other hyper-parameters were tuned outside of the communication scope. In particular, the scaling factor for the additional loss, the model size and the learning rate were explored, but varying these did not change how the other factors affected the results.

\subsection{Communication and network configuration}

\subsubsection{Number of nodes and neighbours}
The overall assessment is that the more the better. However there is an important caveat to this in the resulting size of the local training data partition that is associated with each computing node. Collective accuracy levels did not increase past 20 nodes, to the point where using 30 and 50 nodes performed the same but jumping to 100 nodes caused the performance to decrease. Given the experiments were performed on CIFAR-10, which is a relatively small dataset, the observed lack of improvement can be explained with the resulting small amount of training data available to each node. Indeed, as the number of nodes reached 100, signs of over-fitting were noticeable, e.g. training loss on the local training data rapidly collapsed to zero.

Regarding the number of neighbours, at least in the supervised setting, increasing the number of connections in the graph always helped until a fully connected network was obtained and achieved the best performance. This meant that increasing the level of diversity in communications for each of the compute nodes further increased generalization performance.

\subsubsection{Frequency of broadcasting and consumption \label{sec:exchg-outs_results-freq}}
Although multiple scenarios of consumption frequency were explored, the best performing and logical was to allow each node to consume one sample from its local training data queue for each sample consumed from the rest of the network. Effectively, in a 10-node network, each node is consuming on average 50\% of its actual training data from its private share, while the other 50\% is split among the other 9 nodes, therefore 5.6\% of each other nodes’ private share.

The frequency of broadcasting was constructed in a way that allows for an equilibrium in the number of samples available for consumption. In other words, some scenarios could cause individual nodes to receive too much data from the others to the point where extra data received would need to be dealt with, i.e. flush older or ignore more recent. Instead, the focus was on ensuring the broadcasting frequency allows the consumption to be stable.

On the other end, a too high consumption scenario could arise where each node consumes as much from its private share as from each other nodes, making it just like training on the full dataset. When considering all samples consumed, this would make a single node consume more from others by a factor equal to the number of neighbours. Conceptually, however, this doesn't align well with the human analogy previously described. Therefore, consumption probabilities were set to $0.5$ for all nodes.

\subsubsection{Content of communication}
If the information exchanged between compute nodes are logits, better performance results from lowering the temperature, i.e. hardened outputs. An intuitive extension to sharing logits with low temperature is to share the actual ground truth labels. Although in slight contradiction with the findings in \cite{hinton2015distilling}, these results are confirmed by the observation that exchanging the true labels of the training data performs even better.

The fact that exchanging the true labels performs better than close to zero temperature can be easily explained by the fact that some models make prediction errors on the training set. In other words, with a small temperature that reproduces hardened outputs, a node still has a chance to make an error on the prediction, while the true label is always correct. When making a prediction error, that node sends an incorrect target to another node, and it will negatively affect performance compared to sending the actual true label. Sharing logits later in training was briefly experimented but no significant difference was noticeable.

Broadcasting the top-level features of each compute node performed very badly when compared to the other information communicated.

\subsection{Supervised learning}

The tables \ref{tab:exchg-outs_supervised-results_1} and \ref{tab:exchg-outs_supervised-results_2} show the performance of a single node as well as 10 nodes based on changing the outputs communicated between the compute nodes (logits, true labels, features). Also included in the tables are results for the 20 nodes exchanging true labels as it was the best-performing configuration. All the models in the tables have the same size and are trained with Adam \cite{kingma2014adam} with default learning rates. Both average accuracy of the whole group and collective decision accuracy are denoted as either \textit{Avg acc.} or \textit{Coll. decision} in the tables, respectively. The results further reflect all the previously mentioned configuration and the best of each setting, in particular fully connected communications throughout the network of nodes.

\begin{table}[htbp]
	 \captionsetup{width=.8\textwidth}
    \begin{minipage}{0.8\textwidth}
        \centering
        \begin{tabular}{l | r r}
            \textit{Configuration} & Avg acc. & Coll. decision \\
            \hline
            1 node -- \textit{No sharing} & 84.5 & N.A. \\
            10 nodes -- \textit{No sharing} & 71.1 & 72.9 \\
            10 nodes -- Sharing logits & 79.0 & 84.8 \\
            10 nodes -- Sharing true labels & 81.8 & 86.8 \\
            10 nodes -- Sharing features & 72.0 & 80.1 \\
            20 nodes -- Sharing true labels & -- & \textbf{87.7}  \\
            \hline
        \end{tabular}
        \caption{Validation accuracy (\%) after 5,000 training steps}
        \label{tab:exchg-outs_supervised-results_1}
    \end{minipage}
    \hfill
    \begin{minipage}{0.8\textwidth}
        \centering
        \begin{tabular}{l | r r}
            \textit{Configuration} & Avg acc. & Coll. decision \\
            \hline
            1 node -- \textit{No sharing} & 5,900 & N.A. \\
            10 nodes -- \textit{No sharing} & >13,000 & >13,000 \\
            10 nodes -- Sharing logits & >13,000 & 5,300 \\
            10 nodes -- Sharing true labels & 12,800 & 3,100 \\
            10 nodes -- Sharing features & >13,000 & >13,000 \\
            20 nodes -- Sharing true labels & -- & \textbf{2,700}  \\
            \hline
        \end{tabular}
        \caption{Training steps until reach 85\% accuracy}
        \label{tab:exchg-outs_supervised-results_2}
    \end{minipage}
\end{table}

The table \ref{tab:exchg-outs_supervised-results_1} showcases the performance that configurations can achieve after 5,000 training steps. These training steps can be considered as wall-time for a parallel implementation. In other words, each of the nodes of the network has trained for 5,000 training steps. Both training on either local data or broadcasted data from other nodes were considered as a training step. It can be noticed that sharing the logits (or \textit{soft-labels}) with collective decision achieves similar performance to a single node operating on its own with all the training data available, with both achieving 84.5\% and 84.8\%. In general, leveraging a unified prediction through collective decision making rather than measuring the average accuracy across all nodes allowed a gain of at least 5\% in all configurations, but this increase in performance is expected when using any ensemble method \cite{breiman1996bagging}. This isn't true, however, for one of the baselines, where all nodes are trained independently. Indeed, only a marginal increase can be noticed using the collective decision. For 10 nodes, sharing the features performed poorly as it only reached 80.1\% accuracy with the collective decision, while sharing the true labels reached 86.8\%, the best of all 10 nodes configurations. 

The overall best performing approach was 20 nodes communicating their true labels from the training data with 87.7\%. However, exchanging true labels is basically the same as a traditional ensemble method, only with a weighted dataset resulting from the communication with other nodes rather than having access to all the data. It can be further seen that any of the approaches with a communication protocol outperforms greatly the isolated network of 10 compute nodes, where after 5,000 steps, the accuracy reaches $71.1\%$ and $72.9\%$ for average accuracy and collective decision, respectively.

As for table \ref{tab:exchg-outs_supervised-results_2}, the intent is to show the potential speedup of these approaches when considering the parallelism in play. Making the 10 nodes share logits did not seem to provide much of an acceleration in training when compared to a single node, with both reaching 85\% accuracy at 5,300 and 5,900 steps, respectively. There was, however, a speedup of over 215\% (5,900 vs 2,700) for reaching 85\% accuracy when considering the 20 nodes sharing true labels vs the single node. Sharing the features and the network of nodes without any communications both did not reach 85\% accuracy and therefore were not shown in the table.

\subsection{Unsupervised learning}

Unfortunately, the low performance of features broadcasting in the supervised setting was translated into this new setting, even when considering both the BiGAN and VAE models. It turns out for the unsupervised learning task, having communications between the nodes impacted negatively performance while non-communicating nodes performed slightly better on a downstream classification task, in particular when using the collective decision approach. Table \ref{tab:exchg-outs_unsupervised-results} details the accuracy of the linear classifier trained on top of the extracted features.

The linear classifiers were fully trained using early stopping on a validation set at different stages of the unsupervised learning. Ultimately, the best performing linear classifier on the validation set throughout training was selected and the test set performance is reported.

\begin{table}[htbp]
        \centering
         \captionsetup{width=.8\textwidth}
        \begin{tabular}{l | r r}
            \textit{Configuration} & Avg acc. & Coll. decision \\
            \hline
            \textbf{VAE} \\
            1 node -- \textit{No sharing} & 41.8 & N.A. \\
            10 nodes -- No communication  & 41.6 & \textbf{45.2} \\
            10 nodes -- Full communication   & 41.5 & 42.7 \\
            \hline
            \textbf{BiGAN} \\
            1 node -- \textit{No sharing} & 46.9 & N.A. \\
            10 nodes -- No communication  & 44.5 & \textbf{49.9} \\
            10 nodes -- Full communication   & 44.3 & 46.2 \\
            \hline
        \end{tabular}
        \caption{Test set accuracy (\%) of linear classifier over learned features.}
        \label{tab:exchg-outs_unsupervised-results}
\end{table}

The results showed for the fully communicated network show worst performance than the network of nodes without any communications. It is to be noted, however, that the collective decision approach performed greatly, even for the non-communicating nodes, which is in line with the observation in supervised learning. All approaches could not beat the single node with the average accuracy of the classifiers, but when combined to make a single prediction, it performed better. In particular, for both VAE and BiGAN, the collective decision in the no-communication setting allowed for at least 3\% increase when compared to the single node. Again, this kind of jump in performance is to be expected when using ensemble methods.

Using the KL-Divergence as the feature training objective for the communicated information seemed to help slightly performance when compared to the single node. Indeed, the collective decision with the communication was able to achieve 42.7\%, which is 0.9\% over the single node, while in the BiGAN setting with mean-squared error, it was 0.7\% under that same baseline.

\section{Discussion and conclusion}
The main observation in the supervised setting to be made is that exchanging the true labels performs better than letting the nodes communicate outputs such as the logits or top-level features by a significant margin. As for the unsupervised setting, it was clear that sharing the features seemed to only impact negatively performance. There are, however, still potential uses for the proposed approach to be made in some specific circumstances or setups.

In a scenario where the true label would not be available to be exchanged, e.g. either semi-supervised setting or lost partial data, it was shown that having the compute nodes broadcast their prediction on their data could make the whole group achieve similar performance to what a single model could achieve. This is especially true in a situation where the predictions are known to be good. These results could be interpolated to a scenario where having a single model is simply not feasible and instead of leveraging a distributed SGD implementation that requires the models to exchange the large number of parameter gradients, it could simply exchange the logits between them.

Although exchanging true labels did not provide a linear acceleration with the number of nodes, in very large networks of compute nodes of over 256, it was shown in \cite{anil2018co-distillation} that distributed SGD implementations did not scale up well. Extending the proposed approach to sharing true labels between a much larger network of nodes could still show some speedup. To show the gains of such very large scale network of communicating nodes is left for further work. By further considering why the true labels perform better than logits, this approach could potentially be better if the predictive accuracy of the broadcasting node was better than the consuming node. The better model or node could be replacing the true labels directly and therefore removing the need to broadcast them. An extension of this setting is further described in Chapter \ref{chpt:teach-stud}.

\chapter{Increased Utility Through Selection Of Training Data \label{chpt:teach-stud}}
% Increased Utility Through Selection of Training Data

Training deep learning models with stochastic gradient descent requires randomly selecting samples from the training data. During training of a neural network, it can be anticipated that some samples will be more or less effective in the training of the model and those can be seen as \textit{harder} or \textit{easier}, respectively. In this work, it is proposed to allow a student network, randomly initialized, to communicate with a fully trained network, the teacher, to try and leverage the latter's expertise by instructing the student about which of the training data are \textit{harder}. To identify the difficult examples, rather than simply sending away model outputs or labels as in Chapter \ref{chpt:exchg-outs}, the teacher considers the predictions from the student to evaluate which of the training samples are good training candidates. It is demonstrated that by measuring the \textit{distance} between the predictions of the teacher and the student, it can be used as a proxy of \textit{difficulty} to select samples for the student and therefore accelerate training when compared to randomly sampling training data. This is done by leveraging previous results from Chapter \ref{chpt:exchg-outs} in addition to \cite{hinton2015distilling}'s work on distillation. Furthermore, it will be shown that using the teacher predictions as training targets for the student can further increase convergence speed.

\section{Introduction}

Graduate students often seek shortcuts when studying for a final exam. Instead of going through all of the content, they wish to optimize their grade while not having to go over all the course content. They sometimes do so by their high level of laziness, but more often these shortcuts are taken because the student already understands well a given section of the course material. The student can therefore afford to skip some exercises listed for a chapter, as he trusts his understanding developed through previous experience. Another way a graduate student accelerates his \textit{training} is by leveraging a professor's role and accessibility. He can reach out to a professor who uses his experience to recommend either exercises or additional readings in a way that is beneficial for the student's learning.

By maintaining this analogy of a graduate student, a randomly initialized neural network and its traditional stochastic gradient descent training can be seen as a highly inefficient training procedure. For that randomly initialized neural network, the \textit{student}, going through all training samples certainly has some inefficiencies since it might already have mastered the content of the dataset associated with that sample.

In this work, it is proposed to address these inefficiencies by maintaining a communication channel between the student network and a previously trained teacher network. This communication channel will not only be used to transmit information but also instructions regarding which training samples can accelerate performance. One of the findings leveraged from previous work in Chapter \ref{chpt:exchg-outs} was that sharing the true labels resulted in better performance than sharing the hardened soft-labels as supervised learning targets. A possible explanation to this observation was that the model sending the soft-labels was simply not accurate enough to use its predictions as the ground truth labels. This is where leveraging a teacher and student analogy might prove itself useful.

The teacher will identify which examples can accelerate performance by using the predictions of the student with different measures to quantify the difference to its own predictions. The effect of varying the size of the \textit{database} from which the teacher selects which samples to train the student on will also be detailed. In this setup, the teacher has the possibility of either sending the \textit{normal} training data and the ground truth labels or it can send its own set of predictions as the labels. The communication channel therefore includes both the student sending its prediction to a teacher and receiving the samples along with the appropriate target to use in its training procedure.

A successful acceleration in this simplified setup could be useful in a setup where training the student on the original data is no longer feasible. In a sense, this could be used in semi-supervised learning tasks on labels where the ground truth label is not available for part of the training data.

\section{Method}

\subsection{Teacher/Student framework}

There are two major components of the analogy teacher and student to consider. The first is to consider that a teacher already knows the content of a course, or in the supervised setting, it has a low generalization error. The second aspect to consider is much like the teacher and student relationship in an academic setting, the teacher has the ability to evaluate the student's weaknesses and customize its \textit{training}. More commonly, with its evaluation of the student's skills, he is able to identify which chapters or exercises are the most beneficial for the student to learn the course content.

To incorporate these two components in a supervised learning experimental framework, a model referred to as the teacher was trained until convergence with early stopping, by monitoring its prediction accuracy on a validation set. A second model, the student, is then randomly initialized and its training procedure begins. Throughout training of the student, it will have access to the same data as the teacher, but in addition, it will be able to leverage predictions of the teacher as targets. More details regarding the student's training are detailed in section \ref{sec:teach-stud_student-target}.

\subsection{Data and experimental setup}
\subsubsection{Dataset}
A widely used dataset throughout the machine learning community that was used in this work is the MNIST dataset \cite{lecun1998mnist}. It consists of 60,000 greyscale images of digits between 0 to 9 as well as an additional 10,000 images as the test set. Each of the images is 28x28 pixels which will be used as a single row of 784 pixels. This dataset allows for a simplified experimental setup and does not require a specific model size or architecture in order to show significant performance. Given the objective of this work is to show the acceleration of training, an interesting aspect of using this dataset was the efficiency of training with the MNIST dataset.

From the 60,000 training data samples, 10,000 were set aside to be used as validation set in order to monitor the performance of the teacher for early stopping, as well as training speed for the student.

\subsubsection{Model architecture}
Both the teacher model and the student were constructed as networks with 2 fully connected hidden layers with ReLU \cite{nair2010rectified} in addition to a softmax output layer for each of the 10 classes in the MNIST dataset. The teacher was purposefully set up to have many more hidden units at each layer than the student's network, with 1,200 and 32 respectively. In addition, each hidden layer of the teacher was regularized using dropout \cite{srivastava2014dropout}, while the student was not. The structure of both of these networks was intended to allow the teacher network to acquire more knowledge through its greater capacity than the student model.

For both models, Adam \cite{kingma2014adam} optimization with the default hyper-parameters was used and both the student and teacher architecture were kept the same throughout the experiments. Also, the parameters of the teacher network were trained and kept the same across all the sets of the experiments in order to compare the different effect of student training.

\subsection{Identifying difficult examples for the student \label{sec:teach-stud_student-target}}

The interactions that are further described between the teacher and student models in the proposed framework can be seen analogous as a student answering quizzes and sending them to the teacher. In addition, rather than having the teacher provide feedback on all the submitted quizzes, he only provides the feedback on some of them. This section covers how the teacher selects which samples will be used to train the student, in addition to how they will be used. 

\subsubsection{Evaluating the difficulty}

Let's assume the teacher has a set of training examples from which it needs to select which one will be the most beneficial for the student and let's further assume the ground truth labels are not available. What is proposed here is to leverage the predictions that are made from the student in comparison to the ones made by the teacher. In a way, considering the teacher's prediction as a proxy of the real labels, it can be identified quite easily if the student is wrong. The goal of using them as a proxy is to be able to identify which training samples are more difficult for the student, without requiring the ground truth labels.

However, the intent is to leverage the totality of the probabilities associated with each of the class predictions from the student. There is much more information that can be gathered by considering the full predictive distribution than just its most likely outcome.

It is therefore proposed to measure the difference between both the teacher's and student's predictive distributions as a proxy for difficulty. For a random variable $Y$, the class label, and given probability distributions $P_Y$ and $Q_Y$, the teacher and student's predictive distribution for a given sample, respectively, the different metrics explored are the following.

\textbf{Cross-entropy}. The cross-entropy serves as a natural metric to measure the difficulty of the training samples since it is the actual objective of supervised learning when $P_Y$ is the ground truth label. It measures how much the probability distribution $Q_Y$ differs from $P_Y$, where, for example, $P_Y$ is the target. It is defined as,

\begin{equation}
H(P_Y, Q_Y) = - \sum\limits_y P_Y(y) \log Q_Y(y)
\end{equation}

\textbf{Euclidean distance}. Sometimes called the pairwise distance, it can be used to determine the distance between the two distributions by leveraging the vector form of both distributions. Unlike the cross-entropy, using the Euclidean distance will put an equal weight to each of the class labels. Using the previously defined $P_Y$ and $Q_Y$, it can be defined as,
\begin{equation}
D_E (P_Y, Q_Y) = \Big [  \sum\limits_y (P_Y(y) - Q_Y(y))^2 \Big ] ^ {1/2}
\end{equation}

Out of the above-mentioned error measures, only the Euclidean distance can be considered as a true distance since the cross-entropy is not symmetric. However, the term \textit{distance} will be used more loosely throughout this chapter to reflect any of these error measures.\\

For a given set of training samples, such as a minibatch, once the teacher has both its own set of predictions along with the student's, it can compute any of the metrics for each sample. A sample with a higher error will indicate \textit{current} lower performance from the student and therefore be considered as a harder example. Now equipped with the error of each of the training samples, the teacher is able to select the most difficult ones and send them back to the student.

In order to fairly compare performance between a student exchanging with a teacher and one without such communication, there must be some consideration of the computational cost associated with this approach. In particular, the student still needs to compute the predictions on the training samples to be able to communicate them with the teacher. In addition, the cost of computing the distances is left to the teacher, but it could be assumed that both of these models operate in parallel and in addition, calculating this is much cheaper than doing a backward pass from the student.

\subsubsection{Using soft-labels as student targets}

In addition to selecting which samples will be beneficial to the student, it was important to explore what target to provide to the student once these have been selected. This decision comes down to selecting what information or outputs available from the teacher should be used. Following work in Chapter \ref{chpt:exchg-outs}, an intuitive thing to use are the ground truth labels since it proved to perform better. However, it was proposed this observation occurred because the models exchanging soft-labels did not have good generalization performance. Therefore, using the predictions of the teacher as soft-labels will be considered as a possible target for the student. With any of these approaches, no additional training objective is necessary. They can both leverage the same cross-entropy objective, simply replacing the labels with the soft-labels as detailed in equation \ref{eq:cross-entr-full-distr}.

\subsubsection{Implementation details}
Similarly to the analogy of quizzes, the teacher network manages the set of training samples by stacking the multiple minibatches sent by the student, along with their corresponding predictions. The size of the stack was further important as it allows the teacher to select from a larger set of training samples which ones are the most difficult.

For the implementation, a stack referred to as the \textit{teacher stack} was created with a size controlled by the hyper-parameter $n$, where $n$ represents the number of minibatches it can hold. Before pushing the student's data to the stack, the teacher makes its predictions to be later used to compute the distances.

Once the stack is full, the $m$ highest scoring out of the $n*m$ samples are selected by the teacher as the training data to send back to the student, where $m$ is the batch size. Conceptually, the teacher therefore \textit{compresses} $n$ minibatches into a single one for the student to train on. Although the student does a forward pass on the $n * m$ samples, there is a gain of $(n-1)*m$ backward passes. As the size of the stack increases, it is therefore expected to show a greater acceleration, in terms of accuracy per backward passes. The number of forward passes still needs to be considered, so the mentioned acceleration should take it into consideration.

\section{Results}

In order to show any potential advantages with the proposed approach, it is necessary to compare it with the most basic baseline, a student network training by itself without any communication with the teacher and by using the ground truth labels. This baseline is in essence, the same training procedure as the teacher network, or any randomly initialized neural network, the only difference being their model size and regularization, as previously described.

An additional baseline considered is filling up the teacher stack much like other approaches, but rather than making the teacher select any of the training samples with a metric, simply select which ones to send randomly and send its prediction as soft-labels. This allows to test the impact of actively selecting the samples. Throughout the different configurations tested, all student models were trained by monitoring the validation accuracy and were stopped after 20 epochs of no improvement. Batch size was 32, making the stack of the teacher to choose the hardest examples from multiples of 32.

\subsection{Convergence speed}

\subsubsection{Sample selection and defining student targets}

\begin{figure}[h]
    \centering
    \includegraphics[width=0.8\textwidth]{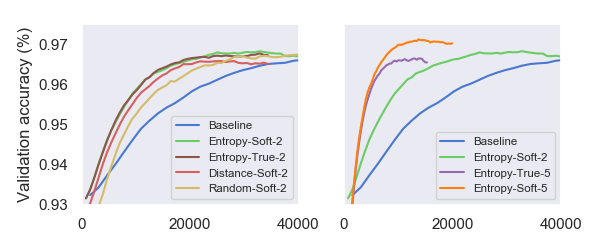}
    \captionsetup{width=.8\textwidth}
    \caption{Validation accuracy (\%) per number of parameter updates of the student network during training. Configurations of the students include a baseline with no communication, (Left) stack size of two minibatches, with both cross-entropy and Euclidean distance as difficulty measure, in addition to comparing soft-labels and true-labels and randomly selection. (Right) The best configuration from stack size two (cross-entropy soft-labels), compared with teacher using a stack size of 5 minibatches with soft-labels and true labels.}
    \label{fig:teacher_student_results}
\end{figure}

The figure \ref{fig:teacher_student_results} shows the validation accuracy during training, and to show the potential acceleration of this approach, by the number of parameter updates, or backward passes, made by the student (in minibatches). 

It can be seen in figure \ref{fig:teacher_student_results} on the left, that all three scenarios using a \textit{distance} and a teacher stack size of 2 minibatches reach about twice as fast the 96\%-97\% mark than the baseline (in blue). Using the Euclidean distance (in red) performs slightly less well than the two other configurations using the cross-entropy between teacher and student predictions as a proxy of difficulty. Considering this, the cross-entropy was selected as the best \textit{distance} to distinguish which examples to send to the student. The baseline of randomly selecting soft-labels to send to the student (in yellow) performs worse than actively selecting based on any of the above mentioned metrics. This can be interpreted as the increased complexity of the communication channel was beneficial for the student's training. Furthermore, using the cross-entropy as a \textit{distance} measure aligns perfectly with the supervised objective. If the teacher's predictions will be used as targets, measuring the difficulty of a sample can now be computed directly with the loss.

In the same figure on the left, it is shown when using a stack size of 2 minibatches, using either the ground truth labels or the teacher's soft-labels did not seem to affect the general performance (in green and brown).

\subsubsection{Teacher stack size}

On the right side of figure \ref{fig:teacher_student_results}, it can be seen that by increasing the size of the teacher stack to 5 minibatches, to make the pool from which the teacher can select the samples bigger, it helps performance. Either with soft or true labels (orange and purple), it greatly reduces the number of backward passes required when compared to both the baseline and using a stack size of 2 minibatches with soft-labels and cross-entropy (in green). Furthermore, using the teacher's prediction as soft-labels for the student's target (in orange) combine with a stack size of 5 minibatches outperforms significantly the student training with directly the true labels. In this work, when the true labels are communicated back to the student as targets, it is important to consider that both student's and teacher's prediction are still used for determining which examples are \textit{harder}. Also, other sizes of \textit{teacher stack} were experimented ranging up to 100 minibatches. However, it did not provide any benefits from scaling and an explanation put forward is the small dataset size. We leave to verify this assumption for future work on much larger networks and datasets. 

\subsection{Generalization performance}

\begin{table}[b]
        \centering
         \captionsetup{width=.8\textwidth}
        \begin{tabular}{l | r r }
            \textit{Configuration} & Accuracy & Backward passes \\
            \hline
            Baseline & 96.4 & 39,000 \\
            \hline
            \textbf{Teacher stack 2 minibatches} \\
            Cross-entropy -- Soft-labels & \textbf{96.9} & 25,000 \\
            Cross-entropy -- True labels & 96.5 & 18,700\\
            Euclidean distance -- Soft-labels & 96.8 & 18,700\\
            Random selection -- Soft-labels & 96.7 & 32,800\\
            \hline
            \textbf{Teacher stack 5 minibatches} \\
            Cross-entropy -- Soft-labels & 96.8 & 13,700\\
            Cross-entropy -- True labels & 96.8 & \textbf{9,100}\\
            \hline
        \end{tabular}
        \caption{Test set accuracy (\%) and number of backward passes for that performance (or parameters updates) of various student configuration based on validation set best scoring parameters. All of the optimization hyper-parameters are the same for the different student networks.}
        \label{tab:teacher_student_test-results}
\end{table}

Although the figure \ref{fig:teacher_student_results} shows convergence speed of the model, it must be considered that peak performance on the validation set cannot be used to compare the generalization performance of different configurations without being biased. To alleviate this, the table \ref{tab:teacher_student_test-results} shows the test set accuracy of the same configurations, using their best model based on the peak validation accuracy parameter's value. These results show that ultimately, all of these approaches have very similar generalization performance. Indeed, all of the approaches are within 0.5\% of the baseline. There is, however, a considerable speedup to get to that performance, in particular when using a bigger teacher stack size. The baseline reached that performance in 39,000 updates/backward passes, while both approaches using a larger \textit{teacher stack} reached it in only 9,100 and 13,700 steps, for true labels and soft-labels, respectively. With this approach applied to a larger dataset, it is anticipated the scaling would be more evident.
 
\section{Conclusion and discussion}

Throughout this work, it was shown that by designing a communication channel used for training instructions between a teacher and a student network, it could allow for accelerated training to reach the same level of generalization performance. Indeed, by allowing the teacher network to compute a \textit{distance} between its own predictions about a given sample and the student's, it can be used as a proxy of sample difficulty and help it train. Creating a minibatch out of harder examples was shown to accelerate significantly convergence speed.

Through that process, the number of backward passes (or parameter updates) can be efficiently reduced by the teacher instructing the student which data to train on. There were further signs that increasing the pool of data from which the teacher can select difficult samples from further increased the acceleration.

The approach presented implies the student has to make a prediction on all datapoints the teacher wishes to consider. The focus was mostly on the fact that even by doing so, there is a speedup because of the gains in backward passes. Other approaches where student predictive ability could be predicted by the teacher would allow to relax that assumption.

Such an approach could provide itself useful in other settings such as where new data is unlabelled but a trained model is available. It could therefore be used to train an additional smaller model without the cost of acquiring labels for those new training samples. Another possible extension of this work would be to consider this approach with a bigger network of nodes. Much like work detailed in Chapter \ref{chpt:exchg-outs}, using predictions from different compute nodes that are experts on their own sets of data may show itself simpler and profitable to transfer knowledge of their data partition.

% \addtocontents{toc}{\protect\newpage}
\chapter{Sharing Internal Representation Through Language\label{chpt:share-lang}}
% Sharing Internal Representations Through Language

Language is the key to humans exchanging communications both diversely and imperfectly between each other. This corresponds to the opposite of the communication protocol of training algorithms used in the machine learning community such as synchronous SGD, where it is mandatory that the information communicated is precise. In the latter, the communication requires high bandwidth by the high number of values and their corresponding high level of precision. However, for some reason, humans can communicate how they perceive their highly complex surroundings with a discrete language through a low bandwidth channel. The language we use is specially crafted in order to allow us to communicate and exchange with our peers. It is therefore of interest to study how a similar language could be useful for deep learning models.

Contrary to work previously described in Chapters \ref{chpt:exchg-outs} and \ref{chpt:teach-stud}, rather than directly exchanging model outputs or training samples, a language is purposefully created between models where it will serve as a way to communicate internal representations. To study how effective such a language is, two models are set up and shown variants of the same input and try to better understand the underlying original input. These two models will be able to communicate, with different levels of complexity with a language. The language created is low bandwidth, discrete and trained to have high mutual information with regards to the partial observation of the broadcasting model. It will be demonstrated that allowing agents to exchange information makes it easier for them to perform a classification task based on their own internal representation.

\section{Introduction}

Training algorithms used for training large neural networks on massive datasets/tasks, such as distributed synchronized gradient descent (SGD), do employ some language to communicate. Indeed, such an approach requires first that all models directly communicate their noisy approximation of the gradients to a central system. Following that, they receive another message containing the actual gradients to update their local parameters.

Such a language that shares gradients, although containing very useful information, has the disadvantage of requiring lots of bandwidth simply given the sheer size of the models. Unlike the gradients that are represented continuously by decimals (up to some precision), the language we use daily as humans consists of selecting discrete words. Furthermore, this discrete language that we use both in writing and speech has some very small bandwidth requirements when compared to parameter gradient tensors. Given we can successfully exchange excessively rich and complex concepts through this discrete language with our peers, developing such a language between machine learning models is of interest.

Although the use of a continuous language with less bandwidth than gradients for dialogue could very well convey more information than a discrete one, focusing on a discrete language allows for a more interesting analysis, in particular with the analogy to human language. Moving forward, not without the implementation difficulties that arise from working with discrete sequences, the focus is on the use and development of a discrete language.

To evaluate how successful a model helps another by sending a message, this work uses partial observations derived from a shared input. The objective is to help the other model/agent understand better the underlying \textit{full} observation behind its own partial observation. Different \textit{levels} of communications are explored throughout the different experiments, in particular one-way broadcasting, two-way broadcasting, in addition to allowing for some feedback from the receiver of the message. It will be illustrated how the latter level of complexity can be rewritten as a different objective tying both models' objective together. To discretize the outputs, the Gumbel-Softmax distribution is used to sample the messages where over time, the temperature used is annealed to ensure the samples become discrete. The message generation is trained through unsupervised learning and it is shown  that having communication does improve performance on a classification task using the previously learned features.

\section{Method}

\subsection{Data}

\subsubsection{Partial observation setting}

Let us assume a world with current state $X$ and two agents $A$ and $B$. One important characteristic of the setup is that $X$ is never fully observable by $A$ nor $B$, but can rather be understood as if an \textit{oracle} could observe the current state of the world. In addition, both $A$ and $B$ witness $X$ at the same time, but from different \textit{angles}, which makes them see it differently. The partial observations of the world by agent $A$ and $B$ are therefore denoted $X_A$ and $X_B$, respectively. An analogy can be seen with how different individuals have different perspectives on the shared world that they live in. The partial observation setup used throughout this work was previously described in \cite{bengio2014evolving}.

The goal of this setting is to consider how as humans we can quite easily communicate about our environment, even though we don't necessarily see it the same way as others that we communicate with. Even though we don't see things exactly the same way, we are able to help each other better understand the underlying world we live in.

\subsubsection{Dataset}

To translate the partial observation setting into a machine learning task, a noisy mask can be applied to training data for each agent and therefore generate two different observations of the same original sample. When created, this mask is random, and when applied to an image, it has the effect of \textit{inverting} pixels selected by that mask. Doing so ensure that all agents have different partial observations of the same original state of the world, or at least, the probability of generating two identical masks given a considerable input size is extremely unlikely. Throughout the experiments, the noise level was kept at $10\%$ and to create the mask, a Bernouilli distribution was sampled for each pixel with probability matching the noise level.  The mask was sampled before training for each agent and kept fixed, i.e. each agent applies their mask on all the inputs, it doesn't change during training.

\begin{figure}[t]
  \centering
  \captionsetup{width=.8\textwidth}   
  \includegraphics[width=.3\textwidth]{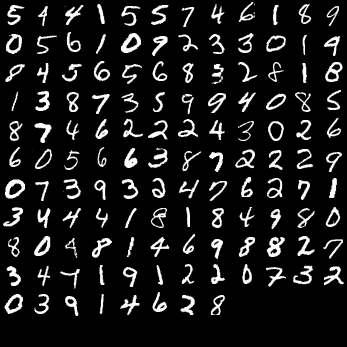}\hfill
  \includegraphics[width=.3\textwidth]{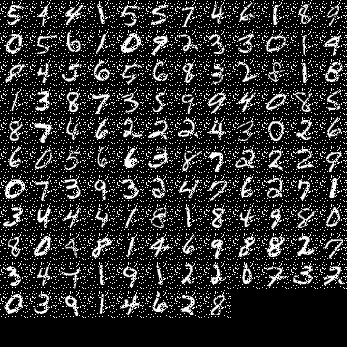}\hfill
  \includegraphics[width=.3\textwidth]{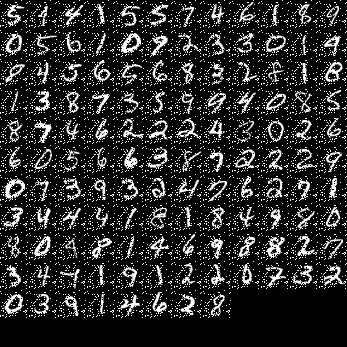}

  \caption{(Left) Original MNIST samples, (Middle \& Right) show the same samples modified by the noisy mask with $10\%$ noise. Each pixel has a probability of $10\%$ to be inverted.}
  \label{fig:noisy-mnist}
\end{figure}

The dataset used in this set of experiments was the MNIST dataset \cite{lecun1998mnist}. It consists of 60,000 images, from which 50,000 are used for training and 10,000 as a validation set, and another 10,000 are given as a test set. Handwritten digits from 0 to 9 represent the 10 possible classes of the dataset. The images are of size 28x28, but are used as a single row of 784 pixels due to model architecture. Some samples along with the resulting partial observations can be seen in Figure \ref{fig:noisy-mnist}.

\subsection{Creating a meaningful language}

\subsubsection{Discretization}
\label{sec:share-lang_discret}

One key component of the human communication is the discretization of internal representations into the discrete language that is used by so many of us. It was therefore important to consider this and build a discrete language.

The language that is explored is to allow models to communicate bits, mainly $0$'s and $1$'s, or rather, a sequence of them. Using bits can be understood as using a language with only a very limited vocabulary size and therefore restricting the number of possible messages. However, by allowing these sequences to be of considerable length, it allows for larger capacity, e.g. for a sequence of only 16 bits, the number of possible messages is $2^{16}$ which equates to 65,536 possibilities.

Generating discrete sequences is known to carry its own set of problems when mixed in the deep learning training procedures because of the backpropagation algorithm. For the gradient descent procedure to work and in particular the chain rule to allow gradients to flow down to the model parameters, all functions must be continuous. However, when making discrete decisions, such as sampling a softmax distribution or taking the argmax of that same distribution, it stops any gradient from flowing back through that operation. In order to leverage gradient optimization, it was necessary to explore alternatives.

To allow the training of the message generation from end-to-end with the main objective, the Gumbel-Softmax \cite{jang2017gumbel} layer was selected. Employing the Gumbel-Softmax distribution to sample the messages allows for the backpropagation to flow through the sampling process, while ensuring, in the limit, that the samples generated are discrete. Any regular training objective usually applied to continuous functions could therefore be used to optimize the messages.

\subsubsection{Vocabulary and message generation}

In order to have the message as an output, different approaches could have been used. At the time of this writing, the current approach used focuses mainly on generating the message \textit{all at once} rather than character by character. Furthermore, given the decision to consider bits as the vocabulary, generating the full message at once is manageable. Indeed, the highest layer of the message generator can be seen as having $n$ heads, where $n$ is the length of the message, and each head has two output units, one for the $0$ and the other for $1$. Each of the output units give a score for that token of the vocabulary and then the Gumbel-Softmax function is applied to these scores to sample a message. In this simplified setup, since only $0$'s and $1$'s can be used as characters, it keeps the size, in number of units, of the output layer pretty reasonable. As for the rest of the message generator model, fully connected hidden layers with rectified linear units were used. 

The other possible approach is to use Recurrent Neural Network to replace this model. With the current state of this work, some successful initial testing was done to ensure the feasibility of the approach with an RNN in combination with the Gumbel-Softmax distribution. However, in particular due to the training time and given the small size of the vocabulary, there is not much gain to be made from moving to this type of model. However for future work with larger vocabulary size, it will be necessary to employ some recurrent connections in the message generator network. For future work, using a RNN could also provide itself useful to allow messages of different lengths. Indeed, it could be used as a way for the receiving model to handle different length messages from other models, or even to generate different length messages to an array of models.

\begin{figure}[t]
	 \captionsetup{width=.8\textwidth}   
     
     \[ \left( \begin{array}{ccccc}
0.54 & 0.47 & 0.52 & 0.57 & 0.39 \\
0.46 & 0.53 & 0.48 & 0.43 & 0.61
\end{array} \right)    \left( \begin{array}{ccccc}
0.94 & 0.12 & 0.80 & 0.99 & 0.00 \\
0.06 & 0.88 & 0.20 & 0.01 & 1.00
\end{array} \right) \] 

    \caption{Vector representation of a message generated of length $5$ with vocabulary size of $2$, and temperature of Gumbel-Softmax set at $4$ (left) and $0.25$ (right). For illustrative purposes, in this figure, the logits and the underlying sample from the Gumbel distribution are the same for the two temperatures.}
\label{fig:share-lang_message-high-temp}
\end{figure}

Given the use of the Gumbel-Softmax layer as the final output layer, the message generated depends on the temperature used to smooth or harden the samples. As an example of a possible message sent from one model to another, see Figure \ref{fig:share-lang_message-high-temp} for the vector representation. The left corresponding to early in training with a high temperature which results in smoother samples, and if applicable, allowing for a greater gradient to flow through. On the right, the same input but as the temperature is annealed throughout training, the same logits generate a \textit{more discrete} message. The vector representation of the message can be seen as a softened one-hot representation. If applicable, training would be done with the softened version of the message, while testing would be done with the one-hot version of the message. In particular, throughout training, the reconstruction of the images by using the hard version of the message was successfully used as a way of ensuring appropriate training.

The algorithm \ref{alg:share-lang} details the steps to generate both the partial observations but also the messages in the previously described setup with two agents, $A$ and $B$.

\begin{algorithm}
    \caption{Partial observation and message generation pseudo-code}
    \label{alg:share-lang}
    \begin{algorithmic}[1] % The number tells where the line numbering should start
    \State Generate masks $m_A$ and $m_B$ based on the noise hyper-parameter
    \State Initialize message generator networks $f_{\theta_A}$ and $f_{\theta_B}$
    
    \While{Training}
      	\State Sample minibatch of data from $D_{train}$
    	\State $X_{A} \leftarrow $ Invert pixels in minibatch based on $m_A$
        \State $X_{B} \leftarrow $ Invert pixels in minibatch based on $m_B$

        \State $s_A \leftarrow f_{\theta_A}(X_A)$ \Comment{generate message from agent A}
        \State $s_B \leftarrow f_{\theta_B}(X_B)$ \Comment{generate message from agent B}
        
        \If{Communication}
    		\State $A$ sends $s_A$ to $B$
        	\State $B$ sends $s_B$ to $A$
    	\EndIf
    \EndWhile

    \end{algorithmic}
\end{algorithm}

\subsubsection{Training objective\label{sec:share-lang_traing-obj}}

One question that arises is how to train or optimize this language. An intuitive answer proposed is that the messages sent should convey as much information possible about the partial observations of the world each agent experiences. Thankfully, these concepts can be tied to probability and information theory quite nicely by considering $X_A$ as the partial observation of agent $A$ and $S_A$ the message it sends to agent $B$ as two random variables. By maximizing the mutual information between these two random variables $I(X_A; S_A)$, it will ensure the message generated is a good replacement of the partial observation. If well trained, the message will be a low bandwidth and informative representation of an agent's partial observation. This objective will serve as the base objective of developing a language without any other task.

Considering both random variables $X_A$ and $S_A$ and their corresponding marginal distributions $p_{X_A}(x_A)$ and $p_{S_A}(s_A)$ as well as the joint distribution $p_{X_A,S_A}(x_A,s_A)$. The mutual information between the partial observation $X_A$ and the message $S_A$ is defined as,

\begin{equation}
I(X_A; S_A) = \mathbb{E}_{X_A,S_A} \log \dfrac{p(x_A, s_A)}{p(x_A) p(s_A)}
\label{eq:lang_mutua}
\end{equation}

To maximize the quantity detailed above, we therefore need to make the joint distribution $p(x_A, s_A)$ differ greatly from the combination of the two marginals $p(x_A)$ and $p(s_A)$. In other words, for samples $x_A$ and $s_A$, the probability associated with that joint observation needs to be high, while the product of both probabilities from their corresponding marginal must be small. This relates to ensuring $X_A$ and $S_A$ are not independent.

To do so, although approaches such as MINE \cite{belghazi2018mutual} could be useful, the traditional GAN objective can be leveraged quite elegantly. In particular, the role of the discriminator in a GAN framework can be viewed as distinguishing between two distributions, i.e. making the true sample and the generated ones far apart. Under that framework, the generator is used to do the opposite and make these two distributions closer, mainly controlling the generated samples’ distribution. In our case, the two \textit{distributions} we wish to make distinguishable is the joint $p(x_A, s_A)$ and the combination of the two marginals, $p(x_A)$ and $p(s_A)$. By making them \textit{far apart}, it increases the mutual information. Similarly to what was done in \cite{brakel2017learning}, it is proposed to use the discriminator's objective of a \textit{vanilla-GAN} to achieve this. Additionally, the generator is trained to further separate the two distributions rather than closer. Doing so, the mutual information between $X_A$ and $S_A$ can be maximized, where \cite{brakel2017learning} minimized it, because $p(s_A)$ and $p(s_A | x_A)$ are controlled by a generator model.

In addition, contrarily to the GAN framework where both the generator and the discriminator are \textit{competing}, our objectives are to generate $S_A$ and make it incorporate information from $X_A$. This therefore includes the generator in the optimization and makes it a \textit{max-max} problem as opposed to a \textit{min-max} problem. The former, based on previous experience, is much easier and stable to train.

Implementation-wise, similarly to the procedure proposed in the MINE framework \cite{belghazi2018mutual}, a pair of $x_A$ and its corresponding (from the joint distribution) $s_A$ are considered as \textit{true} samples in the GAN framework. In addition, a new $x_A'$ is used to generate its corresponding $s_A'$. However the previous $x_A$ is paired with $s_A'$ to form the \textit{fake samples}. The fake samples represent two samples from the two marginal distributions, while the true samples are from the joint distribution. Both $(x_A, s_A)$ and $(x_A, s_A')$ are fed to the discriminator as both the \textit{true samples} and the \textit{fake samples}, respectively.

\subsubsection{Communicating the message with another model}

One of the purposes of developing an informative message is to communicate it to another agent or model. Hopefully, this message will be informative such that it will allow the agent on the receiving end to understand better its own observation.

Considering the two agent setup, one of them is known as the \textit{teacher}, agent A, while the other is the \textit{student}, agent B. The idea here is to develop an objective that ties both agents into a global objective. Without any communications between them, a baseline can be defined where each agent has only the training objective to have their message generation be informative as described in \ref{sec:share-lang_traing-obj}.

In order to add a communication objective, consider the desire for $A$ to have $S_A$, its message, such that it provides few or no additional information than if we had known $X_A$, its partial observation, i.e. $S_A$ has all the information about $X_A$. In addition, $A$ should hope that its message conveys lots of information regarding $B$'s partial observation $X_B$. Using information theory, the former quantity is the conditional entropy of $S_A$ given $X_A$, or $H(S_A | X_A)$, while the latter is known as the mutual information between $X_B$ and $S_A$, or $I(S_A;X_B)$.

In other words, the message from $A$ is \textit{relatable} for $B$, while still having lots of information about its own observation. Putting these two concepts together, a global objective to maximize that ties both agents together can be written as the following,
\begin{equation}
I(S_A; X_B) - H(S_A|X_A)
\label{eq:mutal-info-minus-entropy}
\end{equation}

Interestingly, equation \ref{eq:mutal-info-minus-entropy} can be decomposed and rewritten to obtain the original training objective defined in \ref{sec:share-lang_traing-obj} along with an additional term. Indeed, by expanding the mutual information term, it can be rewritten as,
\begin{equation}
H(S_A) - H(S_A|X_B) - H(S_A | X_A) = I(S_A;X_A) - H(S_A|X_B)
\end{equation}

And finally, the conditional entropy term can be rewritten as,
\begin{equation}
    H(S_A|X_B) = -\sum_{s_A\in S_A,x_b\in X_B} p(s_A, x_B) \log p(s_a | x_b) = \mathbb{E}_{S_A,X_B}\big[-\log p(s_A|x_b)\big]
\end{equation}

Reassembling all the components, we get,
\begin{equation}
I(S_A, X_A) + \mathbb{E}_{S_A,X_B}\big[\log p(s_A|x_b)\big] = R_{info} - L_{likelihood}
\label{eq:share-lang-global-obj}
\end{equation}

The firm term $R_{info}$ in equation \ref{eq:share-lang-global-obj} corresponds to the training objective originally described in section \ref{sec:share-lang_traing-obj}, while the second term $L_{likelihood}$ is actually a likelihood objective on the receiving end of the messages. In other words, agent $B$ tries to predict $s_A|X_B$, which corresponds to the cross-entropy loss. For the implementation, a hyper-parameter was added to control the importance of the likelihood loss in the global objective.

The objective on the receiving end can be analogous to how we try to build a model of the people with whom we communicate. Given our own individual partial observation, we increase the likelihood of a message from our language model based on what the others say on their partial observation. Of course, for this analogy to make sense, both partial observations need to be related in some way, which is the case in the partial observation setup described earlier.

\subsubsection{Communication levels}

To test the effectiveness of the new communication between models, it is proposed to study different \textit{levels} of communication. Each of them will be evaluated by having the agents train a linear classifier on top of the last hidden layer of features before the message output layer. The training of this linear classifier is separated from the main training objective, and only applies to the classification task. The levels of communications will refer to the different gradient sources for the objective defined in equation \ref{eq:share-lang-global-obj}.

For simplicity, let's consider the point of view of the student, or agent B. As previously mentioned, the baseline consists of having the student train only its message generation based on the maximization of the mutual information. The first level of communication consists of having the student reproduce the message the teacher generated. This way, similarly to work done in Chapters \ref{chpt:exchg-outs} and \ref{chpt:teach-stud}, a cross-entropy supervised learning objective is added for the message received.

More formally, let's consider $s_A = [s_A^1, s_A^2, \dots, s_A^T]$ as the message generated by the teacher based on its observation of $X_A$ from $f_{\theta_A}(X_A)$ and $s_B = [s_B^1, s_B^2, \dots, s_B^T]$ the message generated by the student based on its own observation of $X_B$ from $f_{\theta_B}(X_B)$. From this initial level of communication, the gradients from the likelihood term for the student can be computed as,

\begin{equation}
\nabla_{\theta_B} L_{likelihood} =  - \nabla_{\theta_B} \Big[  \sum\limits_{t=1}^T s_A^t \log s_B^t \Big]
\label{eq:share-lang_loss-student-receiving-only}
\end{equation}

Under the formulation above, $s_A^t$ and $s_B^t$ are the real values associated to the probability of activating the bit $t$ in the message sampled from the Gumbel distribution. For more information, refer to section \ref{sec:share-lang_discret} and figure \ref{fig:share-lang_message-high-temp}.

By adding the objective $\mathbb{E}_{S_A,X_B}\big[\log p(s_A|x_b)\big]$, since the expectation is over both $S_A$ and $X_B$, it means some gradients should flow back to the model that generated $S_A$, which in this case, is the teacher or agent $A$ with $f_{\theta_A}$. However, for this \textit{level} of communication, no gradients are propagated back to the broadcaster of the message. This configuration is referred to as the student is only \textit{receiving} the message.

The following level is to allow the gradients from the loss computed in equation \ref{eq:share-lang_loss-student-receiving-only} to flow back to the language generation model on the teacher side $f_{\theta_A}$. This means the teacher will receive, in addition to maximizing $I(S_A, X_A)$, a loss from the likelihood term of the student. This communication has an effect on the student only by allowing the teacher to modify its message generation based on the feedback. It doesn't directly affect the student's message generation model $f_{\theta_B}$. This can be seen as the student sending feedback to the teacher based on its understanding, and similarly, the gradients for the teacher can be computed as,

\begin{equation}
\nabla_{\theta_A} L_{likelihood} = - \nabla_{\theta_A} \Big[ \sum\limits_{t=1}^T s_A^t \log s_B^t \Big]
\label{eq:share-lang_loss-teacher-receiving-only}
\end{equation}

Finally, the last level of communication considered is to further allow the teacher to reproduce the student's language, in addition to allowing gradients to flow back to the student, i.e. allow both of them to communicate and \textit{copy} each other. In other words, both student and teacher are sending their messages while receiving messages from the other. Then, both of them compute their cross-entropy loss using the received message as the target and their own generated message as the prediction. Their cross-entropy loss is then sent back to the sender of the message, and the gradients flow back into the message generation network of the sender.

Considering the total loss computed by the teacher and the student, the gradients from the likelihood terms for the student then become,

\begin{equation}
\nabla_{\theta_B} L_{likelihood}  = - \nabla_{\theta_B} \Big[  \sum\limits_{t=1}^T (s_A^t \log s_B^t +  s_B^t \log s_A^t) \Big]
\label{eq:share-lang_loss-student-both}
\end{equation}

To compare all of these approaches, the classification performance of the student will be monitored throughout training. Conceptually, the different \textit{levels} of communication represent how a model performs when it is either trained alone, on the receiving end of a communication or communicating and receiving feedback from that communication.
\newpage
\section{Results}

The different \textit{levels} of communication explained in the previous section can be analogous to an increased complexity of the communication protocol between two agents. It is therefore interesting to compare how increasing the complexity of that communication affects performance, in particular in comparison to a scenario where the two agents do not communicate.

\begin{figure}[h]
    \centering
    \includegraphics[width=0.6\textwidth]{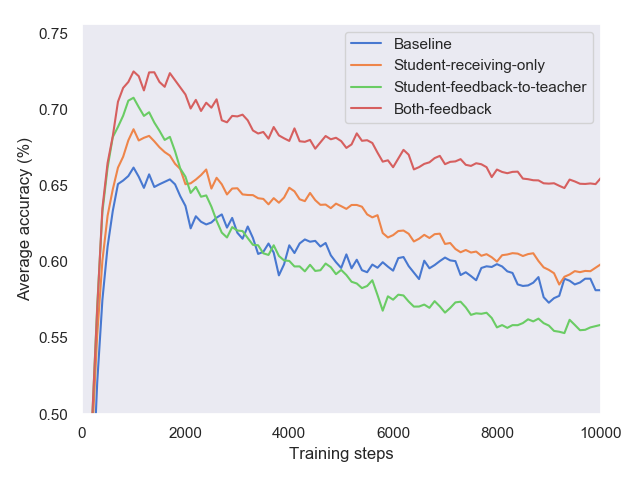}
    \captionsetup{width=.8\textwidth}
    \caption{Validation accuracy (\%) per training steps of the student network during training based on different levels of communication. Results show average of five runs with a noise level of $10\%$. One way communication configuration (orange) has weight of 0.01 to the cross-entropy term, while the two-way communications (green and red) both have 0.005.}
    \label{fig:share-lang-results}
\end{figure}

Figure \ref{fig:share-lang-results} shows performance over the validation set during training of the different \textit{levels} of communication with the length of the message at 32 characters with vocabulary size of 2 (exchanging bits). Both the teacher and the student network were trained at the same time and each of them had their noisy mask kept the same for all the configurations, with a noise level of $10\%$. Furthermore, the performance on the test set of the model parameters according to the best validation set accuracy is shown in table \ref{tab:share-lang-test-results}. The temperature is set to $4.0$ at the start of the training and then annealed by a factor of $0.9$ every $100$ training steps, but never brought lower than $0.5$.

\begin{table}[t]
        \centering
         \captionsetup{width=.8\textwidth}
        \begin{tabular}{l | r }
            \textit{Configuration} & Accuracy \\
            \hline
            Baseline & 64.4 \\
            Student-receiving-only & 65.7 \\
            Student-feedback-to-teacher & 69.4 \\
            Both-feedback & \textbf{71.4} \\
			\hline
        \end{tabular}
        \caption{Test set accuracy (\%) of the different communication level approaches. Performance reported is the average of five runs at the highest validation set accuracy for the student.}
        \label{tab:share-lang-test-results}
\end{table}

From the results, in can be noticed for all scenarios the performance is greater from the start followed by an important decrease in performance. A hypothesis put forward is simply the discretization of the messages. As training progresses, the temperature of the Gumbel-Softmax layer is annealed gradually and over time, there is less and less information in the \textit{hardened} message. A possible approach to mitigate this would be to increase the size of the message, allowing for the extra length in it to carry that lost information. This is, however, left for future work.

The baseline approach without any communication and only the mutual information training objective achieved $64.4\%$ performance with a linear classifier on the test set. By allowing the student to add a cross-entropy term on the received teacher message, it increased slightly the student's performance to $65.7\%$ accuracy. More interestingly, by allowing the cross-entropy error term of the student to flow back in the teacher's message generation model (green in \ref{fig:share-lang-results}), there is a considerable jump in test accuracy to $69.4\%$. This can be seen as the teacher is customizing its language to reduce the predictive error of the student. There is, however, a greater decrease in performance on the validation set than with the other approaches. Finally, the scenario where both models try to predict the other model's message and allow the gradients to flow back shows even greater accuracy. Indeed, it allowed the student to reach $71.4\%$ accuracy on the test set. 

Having a fixed temperature was also tried but did not provide any benefits apart from stabilizing slightly the accuracy over training, but still without beating the peak performance of the annealed temperature setting. Given the temperature seems to be causing some issues, future work could focus on removing the Gumbel temperature by moving towards other approaches to deal with discrete sequences such as REINFORCE \cite{williams1992reinforce}.

To mitigate the early peaking in performance, delaying the training of the linear classifier until the messages were more discrete was also tried. However, it did not provide any gain and was not able to reach the same level of accuracy. It did seem to stabilize slightly the performance, but still was not able to match the peak performance of the \textit{no-delay} approach.

\section{Conclusion and discussion}

It is interesting to point out in addition to work in Chapters \ref{chpt:exchg-outs} and \ref{chpt:teach-stud}, explicitly creating a language trained with its own objective seems to show promising results. The increased complexity from the communication allowed for greater performance in a shared partial observation setting. Indeed, the best performing approach was to allow both models to customize their language based on predictability from the other models.

These results show that making the joint distribution of the message and the partial observation of the broadcaster differ greatly from both these marginal distributions can generate meaningful features. In addition, by adding a cross-entropy term on the receiver's end of the message, it was demonstrated that it can further increase performance, in particular by allowing gradients on that loss to flow back to the sender.

Some future work using a RNN to handle the language between the two models is currently being done to further discretize the language used. Some interesting ways to expand this work is to consider the training objective but with a large number of nodes. Previous work with a large number of compute nodes seemed to have low performance when using communication. However, having this explicit training objective tailored to the message rather than automatically derived from the outputs as in previous Chapters might help break the limitations previously noticed.

\chapter*{Conclusion}
% Conclusion

\noindent In this thesis, a study of various communication channels between deep learning models was presented. This was accomplished by viewing the communication channels with different objectives, ranging from low bandwidth outputs of a model to a language crafted and optimized with the sole purpose of being communicated between two models.

It was shown these low bandwidth messages exchanged between compute nodes of a fully decentralized computing network could speedup some of the training. It was pointed out this approach could allow to give birth to an \textit{internet of computing} given some further research in the pooling of knowledge of the compute nodes. In addition, under a simplified setup of the teacher and student type, a teacher could accelerate the student's training by customizing its training procedure. Indeed, by selecting which samples to train on, by considering both the student's and its own set of predictions, selecting the hardest samples to provide to the student as training samples proved itself to increase convergence speed of the generalization error. Finally, using two randomly initialized models that share a partial observation of an input, it was shown that having a purposefully crafted discrete language can lead to better generalization performance on the learned features. Although the language crafted was mentioned being relatively restrictive, some promising results can pave the way for more flexible language, which is key to extending this proposal to a large number of communicating models.

%%%%%%%%%%%%%%%%%%%%%%%%%%%%%%%%%%%%%
%%   BIBLIOGRAPHIE                  %
%%%%%%%%%%%%%%%%%%%%%%%%%%%%%%%%%%%%%
  % Enlever les commentaires de la prochaine commande si vous préférez que le
  % chapitre s'appelle « Références » plutôt que « Bibliographie » (au choix selon le contexte).
%\let\bibname=\refname   

%% Lorsque vous serez prêt à faire afficher votre bibliographie
%% et vos références, enlevez les commandaires des commandes suivantes
%% et donnez le nom de votre fichier .bib à la commande \bibliography{..}
%% (consultez l'exemple au besoin).  Vous pouvez utiliser le style de votre
%% choix.  Le fichier francaissc.bst est inclus avec le gabarit.  Vous pouvez
%% utiliser ce style avec  \bibliographystyle{francaissc}
% 
\bibliographystyle{plain}		    % Le style de la bibliographie. Notons que les extensions ne sont pas données pour ces deux fichiers.
\bibliography{references}		    % La base de données contenant des entrées bibliographiques. Seules celles référencées dans le texte seront ajoutées \`a la bibliographie.

\end{document}